%% file: main.tex
\documentclass{article}
\usepackage{iclr2027_conference,times}
\usepackage[utf8]{inputenc}
\usepackage[T1]{fontenc}
\usepackage{microtype}
\usepackage{graphicx}
\usepackage[hyphens]{url}
\usepackage{amsmath,amssymb}
\usepackage{xcolor}
\usepackage{booktabs,colortbl,array}
\usepackage{algorithm,algorithmic}
\usepackage{newfloat,listings}
\usepackage{caption}
\usepackage{adjustbox}
\usepackage{placeins}
\DeclareCaptionStyle{ruled}{labelfont=normalfont,labelsep=colon,strut=off} 
\floatstyle{ruled}
\newfloat{listing}{tb}{lst}{}
\floatname{listing}{Listing}

\definecolor{PromptBlue}{RGB}{242,248,252}
\definecolor{PromptBlueFrame}{RGB}{86,132,163}
\definecolor{PromptGreen}{RGB}{244,249,246}
\definecolor{PromptGreenFrame}{RGB}{84,137,106}
\definecolor{PromptImageTag}{RGB}{69,91,112}
\definecolor{PromptSearchTag}{RGB}{35,101,166}
\definecolor{PromptInfoTag}{RGB}{172,65,58}
\definecolor{PromptAnswerTag}{RGB}{39,126,91}
\definecolor{PromptThinkTag}{RGB}{111,78,151}
\definecolor{PromptActionTag}{RGB}{178,96,34}
\definecolor{PromptHistoryTag}{RGB}{46,111,78}
\newcommand{\capsmodel}{\textsc{CAPS}}
\newcommand{\promptimage}[1]{\textcolor{PromptImageTag}{\ttfamily\bfseries #1}}
\newcommand{\promptsearch}[1]{\textcolor{PromptSearchTag}{\ttfamily\bfseries #1}}
\newcommand{\promptinfo}[1]{\textcolor{PromptInfoTag}{\ttfamily\bfseries #1}}
\newcommand{\promptanswer}[1]{\textcolor{PromptAnswerTag}{\ttfamily\bfseries #1}}
\newcommand{\promptthink}[1]{\textcolor{PromptThinkTag}{\ttfamily\bfseries #1}}
\newcommand{\promptaction}[1]{\textcolor{PromptActionTag}{\ttfamily\bfseries #1}}
\newcommand{\prompthistory}[1]{\textcolor{PromptHistoryTag}{\ttfamily\bfseries #1}}

\lstdefinestyle{TaskPrompt}{
  basicstyle=\footnotesize\ttfamily,
  numbers=none,
  backgroundcolor=\color{PromptBlue},
  rulecolor=\color{PromptBlueFrame},
  framerule=0.6pt,
  frame=single,
  framesep=5pt,
  linewidth=0.96\textwidth,
  xleftmargin=0.02\textwidth,
  xrightmargin=0pt,
  captionpos=t,
  columns=fullflexible,
  showstringspaces=false,
  breaklines=true,
  escapeinside={(*@}{@*)}
}

\lstdefinestyle{TextPrompt}{
  basicstyle=\footnotesize\ttfamily,
  numbers=none,
  backgroundcolor=\color{PromptGreen},
  rulecolor=\color{PromptGreenFrame},
  framerule=0.6pt,
  frame=single,
  framesep=5pt,
  linewidth=0.96\textwidth,
  xleftmargin=0.02\textwidth,
  xrightmargin=0pt,
  captionpos=t,
  columns=fullflexible,
  showstringspaces=false,
  breaklines=true,
  escapeinside={(*@}{@*)}
}

\lstdefinestyle{JudgePrompt}{
  basicstyle=\footnotesize\ttfamily,
  numbers=none,
  backgroundcolor=\color{PromptGreen},
  rulecolor=\color{PromptGreenFrame},
  framerule=0.6pt,
  frame=single,
  framesep=4pt,
  aboveskip=5pt,
  belowskip=5pt,
  linewidth=0.98\textwidth,
  xleftmargin=0.01\textwidth,
  xrightmargin=0pt,
  captionpos=t,
  columns=fullflexible,
  showstringspaces=false,
  breaklines=true
}
\definecolor{HeaderGray}{RGB}{245,245,245}
\definecolor{BlockGray}{RGB}{250,250,250}
\definecolor{TopPerformance}{RGB}{216,235,242}
\definecolor{SecondPerformance}{RGB}{248,230,205}
\definecolor{DeepGreen}{RGB}{28,120,78}
\newcommand{\bestvis}[1]{\cellcolor{TopPerformance}\textbf{#1}}
\newcommand{\secondvis}[1]{\cellcolor{SecondPerformance}\underline{#1}}
\newcommand{\gaintext}[1]{\textcolor{DeepGreen}{#1}}
\newcommand{\costsaving}[2]{#1\,{\fontsize{6}{7}\selectfont\gaintext{(#2\%$\downarrow$)}}}

\usepackage{hyperref}
\hypersetup{
  hidelinks,
  pdftitle={Reading is not Reasoning: Bridging the Agentic Policy Gap in Vision-Text Compression},
  pdfauthor={Cheng Fan, Junyi Zhou, Tingzhang Luo, RongJian Xu, Qiyanhui Lu, Mingjian Zhu, Hanting Chen, Jianyuan Guo}
}
\title{Reading is not Reasoning: Bridging the Agentic Policy Gap in Vision--Text Compression}
\author{%
\begin{minipage}[t]{\dimexpr\textwidth-2\tabcolsep\relax}
\centering\small\bfseries
Cheng Fan\textsuperscript{1},
Junyi Zhou\textsuperscript{1},
Tingzhang Luo\textsuperscript{1},
RongJian Xu\textsuperscript{1}\\[2pt]
Qiyanhui Lu\textsuperscript{1},
Mingjian Zhu\textsuperscript{2},
Hanting Chen\textsuperscript{2},
Jianyuan Guo\textsuperscript{1,*}\\[5pt]
\normalfont\small
\textsuperscript{1}City University of Hong Kong\\
\textsuperscript{2}Huawei Technologies Ltd.\\[2pt]
\texttt{chengfan2-c@my.cityu.edu.hk}, \texttt{jianyguo@cityu.edu.hk}\\[2pt]
\textsuperscript{*}Corresponding author
\end{minipage}%
}
\iclrfinalcopy

\begin{document}
\maketitle
\fancyhead{}
\fancyhead[L]{Preprint}
\input{body}
\clearpage
\bibliographystyle{iclr2027_conference}
\bibliography{aaai2027}
\clearpage
\appendix
\input{appendix}
\end{document}

%% file: body.tex
\begin{abstract}
    Multi-step language-model agents repeatedly process growing interaction
    histories, leading to substantial context costs. Vision--text compression
    reduces these costs by rendering history as images, but the resulting modality
    shift creates a marked capability gap. Through controlled evaluations of
    history recovery, matched-state decisions, and complete trajectories, we show
    that this gap cannot be explained by OCR quality alone. Visual-history agents
    exhibit systematic drift in action selection, query formulation, stopping, and
    evidence use, revealing an agentic policy gap. We introduce \textbf{CAPS}, a
    two-stage \textbf{C}ross-modal \textbf{A}gentic \textbf{P}olicy
    \textbf{S}elf-distillation framework that uses the same model's stronger
    text-history policy to supervise its visual-history counterpart. Offline
    trajectory self-distillation transfers successful text-policy behavior to
    visual-history inputs, while online policy self-distillation provides dense
    supervision on states visited by the visual-history policy during reinforcement
    learning. On SearchQA, CAPS improves over AgentOCR by 5.0 and 3.4 percentage
points with 3B and 7B backbones, respectively. On full-history
ALFWorld, the corresponding gains are 15.6 and 14.5 percentage points. Across settings, CAPS reduces average
    memory-context cost by up to 63.3\% and peak cost by up to 83.4\% relative to matched text-history policies. These results show that explicit cross-modal policy self-distillation can preserve agent capability under vision--text
    compression. Our code will be made publicly available.
    \end{abstract}

    \par
    \section{Introduction}
    
    {\small
    \noindent\textit{``Reading furnishes the mind only with materials of knowledge; it is thinking that makes what we read ours.''}\par
    
    \noindent\hfill --- John Locke, \textit{Of the Conduct of the Understanding}\par
    }
    \begin{figure}[!htbp]
        \centering
        \includegraphics[width=0.88\linewidth]{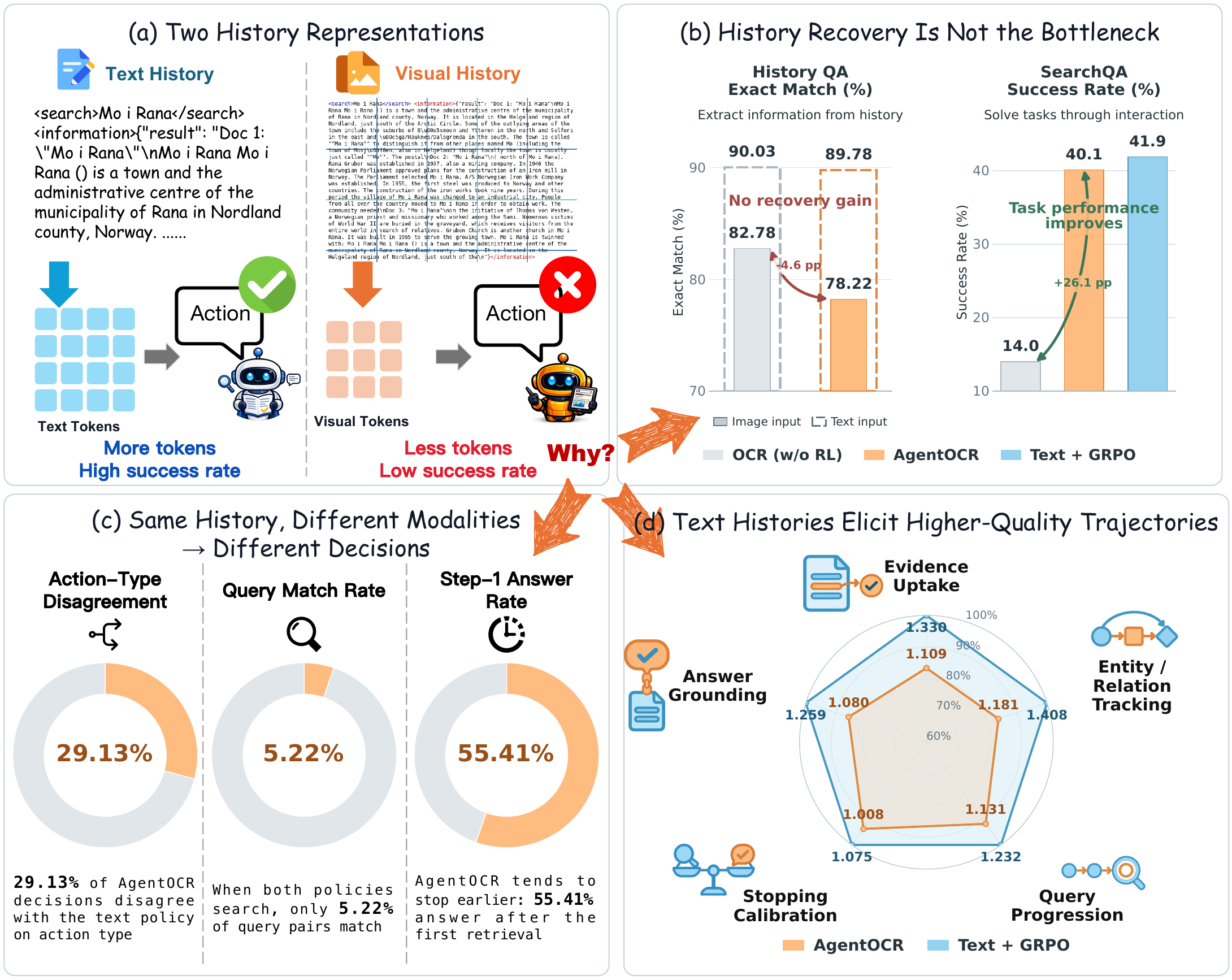}
        
        \caption{{
        \textbf{Textual-history vs. visual-history policy.}
        \textbf{(a)} Two representations of the same interaction history.
        \textbf{(b)} Task improvement is not explained by improved history recovery. Solid/dashed bars evaluate each checkpoint with image/text
        inputs.
        \textbf{(c)} Visual-history policies make different decisions from text-history policies under matched states.
        \textbf{(d)} Text-history policies produce higher-quality trajectories across multiple dimensions.
        }}
        \label{fig:motivation}
        
    \end{figure}
    
    \noindent Large language model agents solve complex tasks by interleaving reasoning and action over multiple rounds of interaction with an environment~\citep{nakano2021webgpt,yao2022react,schick2023toolformer}. Because each decision is conditioned on preceding interactions, new observations and actions are typically appended to the context. As tasks require more retrieval, planning, and tool-use steps, repeatedly processing this growing history incurs increasing token, memory, and latency costs~\citep{kang2025acon,lu2026longseeker,chhikara2025mem0,xu2026mem,zhou2025mem1}. Practical multi-step agents thus need compact histories that preserve information needed for future decisions.\par
    
    Vision--text compression (VTC) offers a promising approach to this trade-off. Rather than discarding or summarizing content, VTC renders text into information-dense images that represent the same content with substantially fewer visual tokens~\citep{wei2025deepseek,wang2024leveraging}.
    AgentOCR~\citep{feng2026agentocr} extends this paradigm to multi-step agents by rendering accumulated observations and actions as visual history and training the agent to act directly from the compressed interface. However, this modality shift incurs a substantial capability loss: despite receiving the same interaction content, the visual-history policy remains markedly weaker than its text-history counterpart, as shown in Fig.~\ref{fig:motivation} (a). Context efficiency, therefore, does not necessarily imply capability preservation. This observation raises a fundamental question: \emph{does this gap arise because the agent cannot reliably read the compressed history, or because it fails to reason and act on what it reads?}\par

    To answer this question, we analyze the gap progressively, from history recovery, through individual decisions, to complete trajectories. On History QA constructed from real trajectories, reinforcement learning greatly improves task success without improving recovery accuracy, indicating that its gains do not come from stronger OCR. Under matched semantic states, changing only the history modality alters action types, search queries, and stopping decisions. At the trajectory level, the visual-history policy also exhibits weaker evidence use, query progression, stopping calibration, and answer grounding. These results reveal an \emph{agentic policy gap} rather than a purely perceptual gap: visual-history agents may read the relevant content yet still fail to reason and act effectively upon it.\par

    Viewing the problem as a policy gap suggests a direct source of
    supervision: the agent's stronger text-history policy. Each interaction
    state provides paired contexts containing the same observations and
    actions, differing only in whether the history is represented as text or
    a rendered image. Because the two policies share the same base model and
    action space, this correspondence enables direct cross-modal behavior
    transfer. We therefore introduce \textbf{CAPS}, a two-stage
    \textbf{C}ross-modal \textbf{A}gentic \textbf{P}olicy
    \textbf{S}elf-distillation framework that transfers reasoning and
    decision behavior from text-history policies to their visual-history counterparts.\par

    CAPS addresses two complementary state distributions. In the first stage,
    \emph{offline trajectory self-distillation} collects successful rollouts from the text-history policy, renders each history prefix into its visual form, and trains the visual-history policy to reproduce the corresponding complete response, including its reasoning and action. This stage transfers successful
    behavior, but its supervision is restricted to states visited by the
    text-history policy. The second stage therefore performs \emph{online policy
    self-distillation} via reinforcement learning. On states induced by the
    visual-history policy's own actions, a frozen text-history policy receives the
    paired textual history and provides full-vocabulary supervision on the
    student-generated response prefixes. CAPS thereby extends cross-modal transfer from successful teacher trajectories to the target policy's own interaction distribution, while retaining environmental reward as the primary learning signal.\par

    Experiments on SearchQA and ALFWorld show that CAPS improves the
    capability--efficiency trade-off across both search and embodied tasks. On SearchQA, CAPS improves over AgentOCR by 5.0 and 3.4 percentage
points with 3B and 7B backbones, respectively. On full-history
ALFWorld, the corresponding gains are 15.6 and 14.5. These gains come with substantially lower context costs than matched text-history policies. Further
    diagnostics show that CAPS improves action and query agreement with the
    text-history policy and trajectory quality across all five
    dimensions.\par

    Our contributions are threefold.
    \textbf{(i)} We provide a controlled diagnosis of the capability loss induced by vision--text compression. The results identify an agentic policy gap that cannot be explained by OCR quality alone.
    \textbf{(ii)} We propose CAPS, a two-stage cross-modal policy self-distillation framework that transfers text-history behavior to visual-history policies through offline and online distillation.
    \textbf{(iii)} Extensive evaluations show that CAPS improves LLM agents' capability--efficiency trade-off.\par

    \par
    \section{Related Work}
    
    \textbf{Multi-step agents.}
    LLM agents solve open-ended tasks through repeated reasoning, action, and
    environment feedback across software, embodied, and search domains
    \citep{jimenez2024swe,wang2023voyager,shridhar2020alfworld,jin2025search}.
    WebGPT and ReAct established prompting and tool-use paradigms
    \citep{nakano2021webgpt,yao2022react}, while later methods directly optimize
    agent policies with reinforcement learning
    \citep{rafailov2023direct,sheng2024hybridflow,wang2025reinforcement}. GRPO uses
    critic-free group-relative updates, and GiGPO adds episode- and step-level
    credit assignment for long-horizon tasks
    \citep{shao2024deepseekmath,feng2026group}. To control growing interaction
    contexts, ACON and LongSeeker use learned compression or orchestration
    \citep{kang2025acon,lu2026longseeker}, whereas SKILL0 internalizes training-time
    skills to reduce inference-time retrieval.\looseness=-1\par

    \noindent \textbf{Vision--text compression.}
    Vision--text compression represents text as images to exploit the information
    density of visual tokens
    \citep{rust2023language,xing2025see,wang2026multimodal,shi2026memocr}. VIST,
    DeepSeek-OCR, and Glyph develop visual representations for long-context
    compression \citep{xing2026vision,wei2025deepseek,cheng2026glyph}. AgentOCR
    extends this approach to multi-step agents with visual histories, segment
    optical caching, and agent-controlled compression \citep{feng2026agentocr},
    while SKILL0 jointly renders skills and interaction histories
    \citep{lu2026skill0}. Reading, Not Thinking attributes the text-to-image
    modality gap partly to degraded reasoning rather than text recognition alone
    \citep{sun2026reading}.\par

    \noindent\textbf{Distillation.}
    Knowledge distillation transfers teacher behavior beyond hard outcome labels
    \citep{gu2024minillm,xu2025speculative}. On-policy distillation instead
    supervises student-sampled sequences, reducing the distribution mismatch of
    fixed off-policy data \citep{agarwal2024policy,yang2025qwen3,xiao2026mimo}.
    OPD further requires compatible reasoning patterns and teacher information not
    already available to the student \citep{li2026rethinking, yuan2026vision}. OPSD conditions one
    model on privileged teacher and standard student contexts, matching their token
    distributions on student-generated trajectories \citep{zhao2026self}. SDAR
    extends this formulation to multi-turn agents by combining reinforcement
    learning with bounded, gated self-distillation \citep{lu2026self}.\par

    \par
    \section{Diagnosing the Visual-History Policy Gap}
    \label{sec:diagnosis}
    Visual-history agents remain substantially weaker than text-history agents
    despite using the same backbone architecture and parameter scale. This gap may
    arise because rendering makes task-relevant text difficult to recover, or
    because the visual interface alters how the model uses accessible information
    to select subsequent actions. We distinguish these two possibilities by
    examining history recovery, decisions under matched interaction states, and
    trajectory-level decision quality.

    \noindent\textbf{Problem Formulation.}
    \label{sec:problem_formulation}
    We consider a finite-horizon interaction between agent and environment
    $\mathcal{E}$. Let $\mathcal{I}$ denote the task instruction and $o_t$ the
    observation at step $t$. The interaction history at step $t$ is
    \begin{equation}
    h_t=(o_1,a_1,o_2,a_2,\ldots,o_t),
    \label{eq:diagnosis:interaction}
    \end{equation}
    where $a_i$ is the action executed at step $i$. Given
    $(\mathcal{I},h_t)$, the policy samples a textual response
    \begin{equation}
    y_t\sim\pi_{\theta}(\cdot\mid\mathcal{I},h_t),
    \qquad
    a_t=\Gamma(y_t),
    \label{eq:diagnosis:response}
    \end{equation}
    where $y_t$ contains intermediate reasoning and an executable action, and the
    task-specific parser $\Gamma$ extracts the action. The environment executes
    $a_t$ and returns $o_{t+1}$, continuing until the task is completed or the
    interaction limit is reached.
    
    Following AgentOCR, the text history is serialized
    into ordered segments and rendered as images:
    \begin{equation}
    (\ell_{t,1},\ldots,\ell_{t,K_t})
    =
    \operatorname{Split}(h_t),
    \quad
    I_t
    =
    \operatorname{Stack}_{k=1}^{K_t}\mathcal{R}(\ell_{t,k};\psi),
    \label{eq:diagnosis:render}
    \end{equation}
    where $\mathcal{R}$ is a deterministic renderer and $\psi$ specifies the font,
    color scheme, padding, and image bounds. Rendered segments are cached by their
    content and reused when available. The resulting $I_t$ is encoded as visual
    tokens and replaces the text tokens corresponding to $h_t$.

    \noindent\textbf{History Recovery Is Not the Primary Bottleneck.}
    Reinforcement learning substantially improves the visual-history policy. We
    first test whether this gain is explained by improved recovery of text from the
    visual history. Using real AgentOCR trajectories on SearchQA, we construct a
    History QA benchmark of $4{,}000$ balanced visual questions. Each question asks
    the model to extract information explicitly recorded in the history, including
    the latest query, a retrieved document title, keyword presence, or a year. The
    benchmark requires visual localization and text extraction but no complex
    sequential decision making. Full construction details are provided in the
    appendix.\par

    As shown in Fig.~\ref{fig:motivation}(b), GRPO raises the average
    SearchQA EM of the 7B visual-history agent from $14.0\%$ to $40.1\%$, while
    image-based History QA decreases from $82.78\%$ to $78.22\%$. When the same
    diagnostic histories are provided as text, the corresponding scores are
    $90.03\%$ and $89.78\%$. 
    Task performance improves without a corresponding gain in history recovery, suggesting that better text extraction alone does not account for the improvement.\par

    \noindent\textbf{Matched Semantic States Reveal a Decision Gap.}
    We next compare the two policies under matched interaction states. We sample
    $10{,}000$ decision prefixes from real SearchQA trajectories generated by the
    text-history policy. For each prefix, the policies receive the same question
    and interaction content; only the history representation differs. Further
    sampling and evaluation details are in Appendix~\ref{app:diagnostics}.\par

    Figure~\ref{fig:motivation}(c) shows a substantial behavioral shift. AgentOCR
    does not match the text policy's next action type on $29.13\%$ of the sampled
    states. At the first decision after receiving a retrieval observation,
    AgentOCR answers in $55.41\%$ of cases, compared with $29.80\%$ for the text
    policy. Even when both policies choose Search, only $5.22\%$ of query pairs
    reach a token Jaccard similarity of at least $0.8$. Thus, replacing text
    history with its visual representation changes whether the agent continues
    searching, when it stops, and what it searches for.\par

    \noindent\textbf{Text Histories Elicit Higher-Quality Trajectories.}
    To assess whether this behavioral difference has a consistent direction in
    quality, we construct a blind trajectory rubric from real SearchQA validation
    trajectories generated by AgentOCR and the text-history policy. We draw a
    stratified sample of $2{,}800$ instances and obtain $2{,}795$ valid judgments
    from DeepSeek-V4-Flash. Complete trajectories are anonymized and scored from
    $0$ to $2$ along five dimensions: evidence uptake, entity/relation tracking,
    query progression, stopping calibration, and answer grounding. We use
    inverse-probability weights to account for the sampling design. Detailed
    criteria of the judge are provided in the appendix.\par

    As shown in Fig.~\ref{fig:motivation}(d), the text-history policy performs
    better in all five dimensions. It scores $1.330$ versus $1.109$ in evidence
    uptake, $1.408$ versus $1.181$ in entity/relation tracking, $1.232$ versus
    $1.131$ in query progression, $1.075$ versus $1.008$ in stopping calibration,
    and $1.259$ versus $1.080$ in answer grounding. The visual-history policy
    therefore differs from the text-history policy in both its decisions and the
    quality of its reasoning trajectories. We identify this decision-level
    discrepancy, rather than OCR alone, as a primary source of the remaining
    performance gap on agentic tasks.\par

    \par
    \section{Method}
    \label{sec:method}
    
    Vision--text compression renders an agent's accumulated text history as images,
    reducing token usage but changing the interface through which the model
    accesses its interaction history. As shown in Sec.~\ref{sec:diagnosis}, this
    change induces a substantial policy shift and degrades performance on agentic
    tasks. Since the same model produces stronger trajectories when conditioned on
    text histories, we use its text-history policy as a teacher for the
    visual-history policy. \textbf{CAPS} transfers policy capability across the two
    history interfaces through two-stage cross-modal agentic self-distillation.
    Offline trajectory self-distillation first initializes the model to reason and
    act from visual histories. During reinforcement learning, online policy
    self-distillation further transfers the text-history policy to states visited
    by the visual-history policy itself. This approach preserves the token savings
    of visual history while reducing the performance gap relative to the
    text-history policy.\par

    \par
    \begin{figure}[!htbp]
        \centering
        \includegraphics[width=0.95\linewidth]{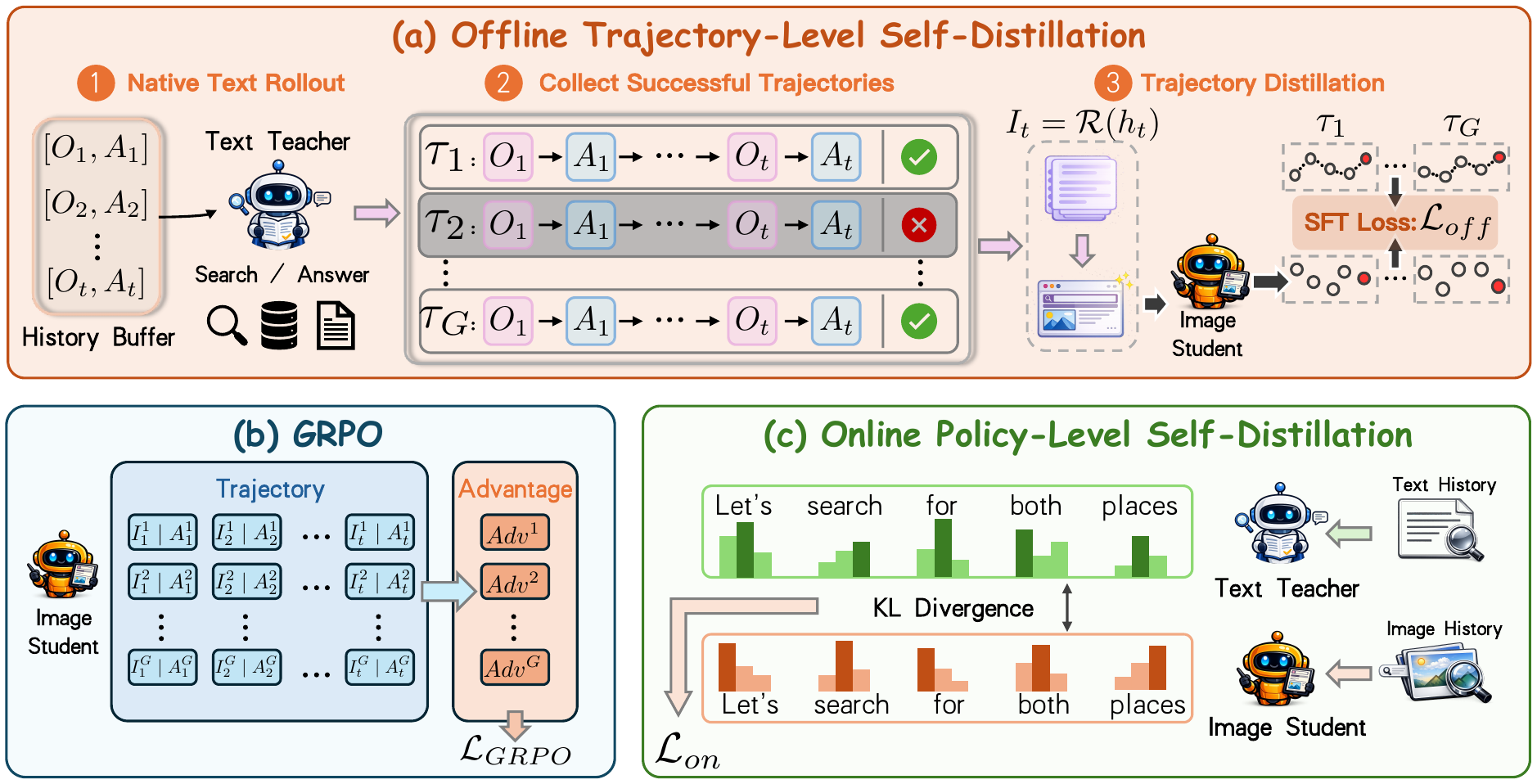}
        
    \caption{{\textbf{Overview of CAPS.}
    \textbf{(a)} Successful text-policy responses supervise the visual-history
    student through $\mathcal{L}_{\mathrm{off}}$.
    \textbf{(b,c)} GRPO and online distillation jointly optimize the student;
    a frozen text teacher provides forward-KL supervision on student-generated
    prefixes at student-visited states.}}
        \label{fig:placeholder}
    
    \end{figure}

    \subsection{Source and Target Policies}
    \label{sec:method:policies}
    
    For each canonical interaction state $(\mathcal{I},h_t)$, the text history and
    its deterministic rendering define a pair of semantically corresponding
    conditioning contexts:
    \begin{equation}
    x_t^{T}=(\mathcal{I},h_t),
    \qquad
    x_t^{I}=(\mathcal{I},I_t).
    \label{eq:method:matched-contexts}
    \end{equation}
    We use a frozen text-history policy $\pi_{\phi}^{T}$ as the source policy and a
    trainable visual-history policy $\pi_{\theta}^{I}$ as the target policy:
    \(\pi_{\phi}^{T}(\cdot\mid x_t^{T})\) and
    \(\pi_{\theta}^{I}(\cdot\mid x_t^{I})\).
    The two policies share the same model architecture and action space and differ
    only in their history representations. We seek to minimize their distributional
    discrepancy over semantically corresponding states while keeping the source
    policy fixed:
    \begin{equation}
    \mathcal{J}_{\mathrm{tr}}(\theta;\phi)
    =
    \mathbb{E}_{(\mathcal{I},h_t)\sim\rho_{\mathrm{tr}}}
    \!\left[
    D\!\left(
    \pi_{\phi}^{T}(\cdot\mid x_t^{T})
    \Vert
    \pi_{\theta}^{I}(\cdot\mid x_t^{I})
    \right)
    \right],
    \quad
    \theta^{\star}
    =
    \arg\min_{\theta}\mathcal{J}_{\mathrm{tr}}(\theta;\phi),
    \quad \phi\text{ fixed},
    \label{eq:method:transfer-objective}
    \end{equation}
    where $D(\cdot\Vert\cdot)$ denotes a discrepancy between response
    distributions and $\rho_{\mathrm{tr}}$ is the interaction-state distribution
    used for cross-modal self-distillation.
    
    \par
    \subsection{Offline Trajectory Self-Distillation}
    \label{sec:method:offline}
    
    \noindent\textbf{Trajectory collection and filtering.}
    Following AgentOCR, we train a text-history agent with
    GRPO and use the resulting policy as the source policy $\pi_{\phi}^{T}$. We
    then collect its interaction trajectories:
    \begin{equation}
    \tau_i=
    \left(
    \mathcal{I}_i,
    \{h_{i,t},y_{i,t}^{T},a_{i,t}\}_{t=1}^{L_i}
    \right).
    \label{eq:method:trajectory}
    \end{equation}
    We retain trajectories that successfully complete the task, contain parsable
    actions, and have responses that are not truncated by the length limit. When
    multiple valid trajectories are available for the same task instance, a
    deterministic ranking selects one representative trajectory, prioritizing
    fewer interaction steps, fewer repeated actions, shorter responses, and
    shorter input contexts. Full data-processing details are provided in the
    appendix.
    
    For each decision step $t$ in a retained trajectory, $y_{i,t}^{T}$ denotes the
    complete source-policy response, including its reasoning and action. We keep
    the task instruction, interaction content, and response unchanged, and replace
    only the history representation:
    \begin{equation}
    (\mathcal{I}_i,h_{i,t},y_{i,t}^{T})
    \longmapsto
    (\mathcal{I}_i,I_{i,t},y_{i,t}^{T}).
    \label{eq:method:offline-pair}
    \end{equation}
    The resulting offline dataset is
    \(
    \mathcal{D}_{\mathrm{off}}
    =
    \{(\mathcal{I}_i,I_{i,t},y_{i,t}^{T})\}_{i,t}
    \).
    
    \noindent\textbf{Training objective.}
    The visual-history policy learns source responses through next-token prediction:
    \begin{equation}
    \begin{aligned}
    \mathcal{L}_{\mathrm{off}}(\theta)
    &=
    -\mathbb{E}_{(\mathcal{I},I,y^{T})\sim\mathcal{D}_{\mathrm{off}}}
    \Biggl[
    \frac{1}{|y^{T}|}
    \sum_{k=1}^{|y^{T}|}
    \log\pi_{\theta}^{I}
    \left(y_k^{T}\mid\mathcal{I},I,y_{<k}^{T}\right)
    \Biggr].
    \end{aligned}
    \label{eq:method:offline-loss}
    \end{equation}
    Full-response supervision distills the teacher's intermediate reasoning and action generation.
    
    \par
    \subsection{Visual-History Reinforcement Learning}
    \label{sec:method:grpo}
    
    The checkpoint obtained from offline trajectory self-distillation initializes the
    visual-history policy, which is then optimized with GRPO
    \citep{feng2026agentocr,lu2026skill0}. For a task $\mathcal{I}$, the policy
    samples a group of $G$ trajectories $\{\tau_i\}_{i=1}^{G}$ with environmental
    returns $\{R_i\}_{i=1}^{G}$. Each trajectory's relative advantage $\widehat{A}_i$ is estimated by normalizing its return within the group.
    The GRPO objective is
    \begin{equation}
    \mathcal{L}_{\mathrm{GRPO}}(\theta)
    =
    -\mathbb{E}\!\left[
    \frac{1}{\sum_{i=1}^{G}|\tau_i|}
    \sum_{i=1}^{G}\sum_{t=1}^{|\tau_i|}
    \min\!\left(
    r_{i,t}(\theta)\widehat{A}_i,\,
    \operatorname{clip}\!\left(
    r_{i,t}(\theta),1-\varepsilon,1+\varepsilon
    \right)\widehat{A}_i
    \right)
    \right],
    \label{eq:method:grpo-loss}
    \end{equation}
    where the importance sampling ratio is
    \begin{equation}
    r_{i,t}(\theta)
    =
    \frac{
    \pi_{\theta}^{I}
    (y_{i,t}\mid\mathcal{I},\tau_{i,<t})
    }{
    \pi_{\theta_{\mathrm{old}}}^{I}
    (y_{i,t}\mid\mathcal{I},\tau_{i,<t})
    }.
    \label{eq:method:ratio}
    \end{equation}
    
    \par
    \subsection{Online Policy Self-Distillation}
    \label{sec:method:online}
    
    Offline trajectory self-distillation covers only states visited by the source
    policy. During reinforcement learning and deployment, however, the visual
    policy visits history states induced by its own previous actions. We therefore
    use online policy distillation~\citep{zhao2026self} as an auxiliary objective
    during GRPO, distilling the source policy on the target policy's state
    distribution.
    For every state visited by the visual policy during rollout, the system retains
    the canonical text history formed by the same student actions and
    environment observations. The student and teacher receive the paired contexts
    $x_t^{I}$ and $x_t^{T}$ from Eq.~\eqref{eq:method:matched-contexts},
    respectively.
    The source policy $\pi_{\phi}^{T}$ is a frozen text-history policy, whereas the
    target policy $\pi_{\theta}^{I}$ is the visual-history policy being optimized. The teacher receives no ground-truth solution and instead
    reads the text-history representation of the same student-visited
    interaction history.
    The student first samples a response $\widehat{y}$ from
    $\pi_{\theta}^{I}(\cdot\mid x_t^{I})$. The teacher and student then compute
    full-vocabulary next-token distributions on the same student-generated prefix
    $\widehat{y}_{<k}$:
    \begin{equation}
    p_k^{S}(v)
    =
    \pi_{\theta}^{I}(v\mid x_t^{I},\widehat{y}_{<k}),
    \quad
    p_k^{T}(v)
    =
    \pi_{\phi}^{T}(v\mid x_t^{T},\widehat{y}_{<k}),
    \quad v\in\mathcal{V}.
    \label{eq:method:online-distributions}
    \end{equation}
    
    \noindent\textbf{Full-vocabulary forward KL.}
    For sample $b$ in an update batch $\mathcal{B}$, the teacher-to-student
    forward KL at response position $k$ is
    \begin{equation}
    D_{b,k}^{\mathrm{FKL}}
    =
    \sum_{v\in\mathcal{V}}
    p_{b,k}^{T}(v)
    \log\frac{p_{b,k}^{T}(v)}{p_{b,k}^{S}(v)}.
    \label{eq:method:online-kl}
    \end{equation}
    To prevent unusually large position-level KL values from dominating the
    distillation gradient, we cap each value at $\tau$. Let
    $m_{b,k}\in\{0,1\}$ denote the intersection of the student and teacher response
    masks. If the complete response contains no valid environment action, all of
    its positions are set to zero. The online distillation objective is
    \begin{equation}
    \mathcal{L}_{\mathrm{on}}(\theta)
    =
    \frac{
    \textstyle\sum_{b\in\mathcal{B}}\sum_k
    m_{b,k}\min\!\left(D_{b,k}^{\mathrm{FKL}},\tau\right)
    }{
    \textstyle\max\!\left(
    1,\sum_{b\in\mathcal{B}}\sum_k m_{b,k}
    \right)
    }.
    \label{eq:method:online-loss}
    \end{equation}
    Here, $\tau$ is the clipping threshold.

    \noindent\textbf{Joint training objective.}
    Environmental reward remains the primary learning signal, while online
    distillation provides dense policy supervision on student-visited states. The
    joint objective is
    \begin{equation}
    \mathcal{L}(\theta)
    =
    \mathcal{L}_{\mathrm{GRPO}}(\theta)
    +\lambda\mathcal{L}_{\mathrm{on}}(\theta),
    \label{eq:method:joint-loss}
    \end{equation}
    where $\lambda$ controls the strength of online policy self-distillation.
    
    \par
    \section{Experiments}
    \label{sec:experiments}
    
    \par
    \subsection{Experimental Setup}
    \label{sec:experiments:setup}
    
    \noindent\textbf{Benchmarks.}
    We evaluate CAPS on two multi-turn agent benchmarks. SearchQA follows the
    evaluation suite used by AgentOCR and SKILL0
    \citep{feng2026agentocr,lu2026skill0}, covering three single-hop datasets (NQ \citep{kwiatkowski2019natural},
    TriviaQA \citep{joshi2017triviaqa}, and PopQA \citep{mallen2023not}) and four multi-hop datasets (HotpotQA \citep{yang2018hotpotqa}, 2Wiki \citep{ho2020constructing}, MuSiQue \citep{trivedi2022musique}, and
    Bamboogle \citep{press2023measuring}). NQ and HotpotQA are used for training, and the remaining datasets
    measure out-of-domain generalization. The agent queries an E5 retriever that
    returns the top three passages and interacts with the environment for at most
    four steps. ALFWorld \citep{shridhar2020alfworld} contains six categories of embodied household tasks and
    requires substantially longer interaction. Following AgentOCR and SKILL0, our
    main comparison retains the full interaction history ($H=50$). We also report
    $H=2$, which follows the default configuration of GiGPO.\par

    \noindent\textbf{Baselines.}
    We compare CAPS with text-history and visual-history baselines. \textit{Text
    (w/o RL)} and \textit{OCR (w/o RL)} use the original text history and the
    rendered visual history, respectively, without RL. \textit{Text + GRPO
    (Qwen2.5)} is the text-history policy reported by AgentOCR \citep{feng2026agentocr}. We additionally
    train \textit{Text + GRPO (Qwen2.5-VL)}, which uses the same VLM family as
    CAPS and serves as its source policy. The visual-history agent baselines are
    AgentOCR and SKILL0, both of which use vision--text compression. We evaluate
    3B and 7B models throughout.\par

    \noindent\textbf{Training details.}
    CAPS uses Qwen2.5-VL-3B-Instruct and Qwen2.5-VL-7B-Instruct as the target
    backbones \citep{bai2025qwen25vltechnicalreport}, with a size-matched Text + GRPO (Qwen2.5-VL) policy as the teacher. The core training configuration and hyperparameters follow
    AgentOCR. Offline trajectory self-distillation uses successful teacher
    trajectories collected from the same training split used by subsequent GRPO;
    it introduces no additional task instances. Full hyperparameters appear in the appendix.\looseness=-1\par

    \par
    \subsection{Main Results}
    \label{sec:experiments:main-results}
    
    \begin{table}[!htbp]
    \centering
    \caption{\textbf{SearchQA performance and memory-context cost.}
    We report EM (\%) and the average and peak memory-context token cost per step.
    $^{\dagger}$Results reported by AgentOCR~\citep{feng2026agentocr}.
    $^{\ddagger}$Results reported by SKILL0~\citep{lu2026skill0}.
    Bold and underlined values denote the best and second-best performance.}
    \label{tab:searchqa-main}
    
    \small
    \setlength{\tabcolsep}{2pt}
    \renewcommand{\arraystretch}{1.12}
    
    \begin{adjustbox}{max width=\linewidth}
    \begin{tabular}{@{}l*{10}{c}@{}}
    \toprule
    \rowcolor{HeaderGray}
    \textbf{Method}
    & \multicolumn{3}{c}{\textbf{Single-Hop}}
    & \multicolumn{4}{c}{\textbf{Multi-Hop}}
    & \textbf{Overall}
    & \multicolumn{2}{c}{\textbf{Tokens/Step}} \\
    \cmidrule(lr){2-4}
    \cmidrule(lr){5-8}
    \cmidrule(lr){9-9}
    \cmidrule(lr){10-11}
    & NQ & TriviaQA & PopQA
    & HotpotQA & 2Wiki & MuSiQue & Bamboogle
    & Avg.$\uparrow$ & Avg.$\downarrow$ & Max.$\downarrow$ \\
    \midrule
    
    \rowcolor{BlockGray}
    \multicolumn{11}{c}{\textit{Qwen2.5-(VL)-3B-Instruct}} \\
    
    Text (w/o RL)$^{\dagger}$
    & 9.4 & 31.3 & 19.8
    & 15.0 & 14.8 & 4.7 & 16.8
    & 15.9 & 0.48k & 7.34k \\
    
    Text + GRPO (Qwen2.5)$^{\dagger}$
    & 39.3 & 60.6 & 41.1
    & 37.4 & 34.6 & 15.4 & 26.4
    & 36.4 & 0.61k & 9.55k \\
    
    Text + GRPO (Qwen2.5-VL)
    & 45.5 & 62.3 & 47.7
    & 42.4 & 42.5 & 16.0 & 35.2
    & 41.7 & 0.57k & 10.63k \\
    
    \midrule
    
    OCR (w/o RL)$^{\dagger}$
    & 10.2 & 27.7 & 10.9
    & 9.1 & 12.2 & 3.7 & 15.2
    & 12.7 & 0.15k & 1.33k \\
    
    AgentOCR$^{\dagger}$
    & 38.6 & 56.5 & 41.7
    & 33.6 & 30.7 & \secondvis{14.6} & 24.0
    & 34.2 & \underline{0.26k} & \underline{2.50k} \\
    
    SKILL0$^{\ddagger}$
    & \secondvis{39.8}
    & \secondvis{57.5}
    & \secondvis{42.3}
    & \secondvis{35.1}
    & \secondvis{33.7}
    & 13.3
    & \bestvis{63.7}
    & \bestvis{40.8}
    & \textbf{0.18k}
    & -- \\
    
    \textbf{CAPS}
    & \bestvis{42.5}
    & \bestvis{60.1}
    & \bestvis{44.4}
    & \bestvis{40.1}
    & \bestvis{37.8}
    & \bestvis{15.4}
    & \secondvis{34.4}
    & \secondvis{39.2}
    & \costsaving{0.37k}{34.8}
    & \costsaving{\textbf{1.76k}}{83.4} \\
    
    \midrule
    
    \rowcolor{BlockGray}
    \multicolumn{11}{c}{\textit{Qwen2.5-(VL)-7B-Instruct}} \\
    
    Text (w/o RL)$^{\dagger}$
    & 10.4 & 32.4 & 22.3
    & 15.8 & 15.4 & 7.2 & 19.2
    & 17.5 & 0.70k & 10.96k \\
    
    Text + GRPO (Qwen2.5)$^{\dagger}$
    & 45.1 & 63.7 & 44.0
    & 43.6 & 43.2 & 16.8 & 37.6
    & 41.9 & 0.73k & 13.84k \\
    
    Text + GRPO (Qwen2.5-VL)
    & 46.9 & 65.5 & 48.5
    & 44.6 & 38.1 & 18.7 & 38.4
    & 43.0 & 0.60k & 3.83k \\
    
    \midrule
    
    OCR (w/o RL)$^{\dagger}$
    & 6.9 & 30.4 & 12.0
    & 10.5 & 9.1 & 5.5 & 24.0
    & 14.0 & 0.26k & 2.21k \\
    
    AgentOCR$^{\dagger}$
    & \secondvis{43.1}
    & 61.0
    & \secondvis{45.4}
    & \secondvis{40.8}
    & \secondvis{38.3}
    & 15.7
    & 36.8
    & 40.1
    & 0.36k
    & \underline{2.65k} \\
    
    SKILL0$^{\ddagger}$
    & 42.7
    & \secondvis{61.1}
    & 45.3
    & 40.0
    & \secondvis{38.3}
    & \secondvis{16.4}
    & \bestvis{66.9}
    & \bestvis{44.4}
    & \underline{0.34k}
    & -- \\
    
    \textbf{CAPS}
    & \bestvis{44.4}
    & \bestvis{64.2}
    & \bestvis{47.1}
    & \bestvis{42.9}
    & \bestvis{42.3}
    & \bestvis{18.3}
    & \secondvis{45.6}
    & \secondvis{43.5}
    & \costsaving{\textbf{0.29k}}{51.3}
    & \costsaving{\textbf{1.74k}}{54.6} \\
    
    \bottomrule
    \end{tabular}
    \end{adjustbox}
    
    \end{table}
    
    \noindent\textbf{SearchQA.}
    Table~\ref{tab:searchqa-main} reports the SearchQA results. CAPS obtains the
    best visual-history result on all six datasets other than Bamboogle at both
    model scales. With the 3B backbone, it reaches 39.2 average EM, improving over
    AgentOCR by 5.0 points. With the 7B backbone, it reaches 43.5, a 3.4-point gain
    over AgentOCR and 0.5 points above its text-history teacher. SKILL0 remains
    first in the overall average because of its substantially higher Bamboogle
    score; CAPS ranks second overall while leading on the other six datasets.
    Relative to the matched text-history teacher, CAPS reduces average and peak
    memory-context cost by 34.8\% and 83.4\% for 3B, and by 51.3\% and 54.6\% for
    7B. Thus, the visual-history policy approaches or exceeds its text-history
    teacher while processing substantially fewer context tokens.\par

    \begin{table}[!htbp]
    \small
    \centering
    \caption{\textbf{ALFWorld performance and memory-context cost with
    Qwen2.5-(VL)-3B/7B-Instruct.}
    We report success rate (\%), the history window $H$, and the average and
    peak memory-context token cost per step.
    $^{\dagger}$Results reported by AgentOCR~\citep{feng2026agentocr}.
    $^{\ddagger}$Results reported by SKILL0~\citep{lu2026skill0}.
    Bold and underlined values denote the best and second-best performance. For our evaluations, Avg. denotes episode success averaged over
validation batches.}
    \label{tab:alfworld-main}
    
    \setlength{\tabcolsep}{2pt}
    \renewcommand{\arraystretch}{1.12}
    
    \begin{adjustbox}{max width=\linewidth}
    \begin{tabular}{@{}l*{10}{c}@{}}
    \toprule
    \rowcolor{HeaderGray}
    \textbf{Method} & \textbf{$H$}
    & \textbf{Pick} & \textbf{Look} & \textbf{Clean}
    & \textbf{Heat} & \textbf{Cool} & \textbf{Pick2}
    & \textbf{Avg.$\uparrow$}
    & \multicolumn{2}{c}{\textbf{Tokens/Step}} \\
    \cmidrule(lr){10-11}
    & & & & & & & & & Avg.$\downarrow$ & Max.$\downarrow$ \\
    \midrule
    
    \rowcolor{BlockGray}
    \multicolumn{11}{c}{\textit{Qwen2.5-(VL)-3B-Instruct}} \\
    
    Text (w/o RL)$^{\dagger}$
    & 50 & 34.7 & 18.4 & 12.7 & 7.3 & 14.5 & 10.4
    & 16.3 & 1.09k & 3.04k \\
    
    Text + GRPO (Qwen2.5)$^{\dagger}$
    & 50 & 92.6 & 85.7 & 70.6 & 86.6 & 79.3 & 65.0
    & 79.9 & 1.02k & 3.13k \\
    
    Text + GRPO (Qwen2.5-VL)
    & 50 & 95.5 & 91.6 & 96.6 & 93.8 & 80.9 & 84.5
    & 91.0 & 0.54k & 2.25k \\
    
    Text + GRPO (Qwen2.5-VL)
    & 2 & 91.1 & 86.2 & 92.8 & 77.1 & 56.6 & 75.2
    & 80.5 & \underline{0.09k} & \underline{0.46k} \\
    
    \midrule
    
    OCR (w/o RL)$^{\dagger}$
    & 50 & 42.8 & 21.8 & 10.1 & 6.2 & 6.2 & 9.9
    & 16.2 & 0.49k & 1.63k \\
    
    AgentOCR$^{\dagger}$
    & 50 & 91.9 & \secondvis{81.8} & 76.0 & 73.3 & 76.1 & 70.0
    & 78.2 & \underline{0.38k} & \underline{1.14k} \\
    
    SKILL0$^{\ddagger}$
    & 50 & \bestvis{95.6} & 80.4 & \bestvis{100.0}
    & \secondvis{86.7} & \secondvis{78.7} & \secondvis{75.2}
    & \secondvis{87.9} & \underline{0.38k} & -- \\
    
    \textbf{CAPS}
    & 50 & \secondvis{94.4} & \bestvis{91.7} & \secondvis{97.2}
    & \bestvis{100.0} & \bestvis{93.0} & \bestvis{83.8}
    & \bestvis{93.8}
    & \costsaving{\textbf{0.20k}}{62.5}
    & \costsaving{\textbf{0.95k}}{57.6} \\
    
    \textbf{CAPS}
    & 2 & 93.9 & 71.7 & 75.0 & 53.1 & 46.9 & 79.5
    & 74.6
    & \costsaving{\textbf{0.03k}}{60.5}
    & \costsaving{\textbf{0.11k}}{75.5} \\
    
    \midrule
    
    \rowcolor{BlockGray}
    \multicolumn{11}{c}{\textit{Qwen2.5-(VL)-7B-Instruct}} \\
    
    Text (w/o RL)$^{\dagger}$
    & 50 & 67.6 & 35.4 & 19.3 & 31.3 & 30.1 & 4.4
    & 31.3 & 1.08k & 3.36k \\
    
    Text + GRPO (Qwen2.5)$^{\dagger}$
    & 50 & 92.6 & 93.8 & 85.2 & 80.0 & 82.7 & 56.5
    & 81.8 & 0.95k & 2.81k \\
    
    Text + GRPO (Qwen2.5-VL)
    & 50 & 100.0 & 100.0 & 93.2 & 96.9 & 83.4 & 82.0
    & 92.6 & 0.58k & 2.90k \\
    
    Text + GRPO (Qwen2.5-VL)
    & 2 & 95.6 & 64.1 & 86.9 & 96.9 & 67.7 & 70.8
    & 83.2 & \underline{0.09k} & \underline{0.69k} \\
    
    \midrule
    
    OCR (w/o RL)$^{\dagger}$
    & 50 & 61.0 & 33.2 & 17.2 & 11.6 & 12.5 & 16.5
    & 25.3 & 0.47k & 1.36k \\
    
    AgentOCR$^{\dagger}$
    & 50 & 95.6 & \secondvis{96.2} & 78.1
    & 73.2 & 72.4 & 72.0
    & 81.2 & 0.43k & \underline{1.22k} \\
    
    SKILL0$^{\ddagger}$
    & 50 & \bestvis{100.0} & 85.8 & \secondvis{94.6}
    & \secondvis{81.9} & \secondvis{85.7} & \secondvis{80.1}
    & \secondvis{89.8} & \underline{0.41k} & -- \\
    
    \textbf{CAPS}
    & 50 & \secondvis{96.9} & \bestvis{96.7} & \bestvis{100.0}
    & \bestvis{92.3} & \bestvis{100.0} & \bestvis{87.7}
    & \bestvis{95.7}
    & \costsaving{\textbf{0.21k}}{63.3}
    & \costsaving{\textbf{0.87k}}{70.0} \\
    
    \textbf{CAPS}
    & 2 & 94.8 & 42.6 & 98.3 & 85.6 & 79.0 & 88.0
    & 84.8
    & \costsaving{\textbf{0.03k}}{63.5}
    & \costsaving{\textbf{0.13k}}{81.6} \\
    
    \bottomrule
    \end{tabular}
    \end{adjustbox}
    
    \end{table}

    \noindent\textbf{ALFWorld.}
    Table~\ref{tab:alfworld-main} shows consistent gains on longer embodied tasks.
    Under the full-history setting ($H=50$), CAPS reaches 93.8\% success with the
    3B backbone and 95.7\% with the 7B backbone. These results exceed SKILL0 by
    5.9 points and AgentOCR by 15.6 and 14.5 points, respectively. CAPS also
    outperforms its text-history teacher by 2.8 points for 3B and 3.1 points for
    7B. The 7B policy reduces average and peak memory-context cost by 63.3\% and
    70.0\%, while the 3B policy reduces them by 62.5\% and 57.6\%. With $H=2$,
    CAPS obtains 74.6\% and 84.8\% success with 3B and 7B, respectively. Without
    modifying the visual-history renderer or explicitly optimizing image
    compression, CAPS uses fewer average context tokens than AgentOCR and SKILL0
    at both model scales. We attribute this gain to better policy optimization:
    more effective reasoning and action selection complete tasks in fewer steps,
    yielding shorter histories.\par

    \par
    \subsection{Ablation Studies}
    \label{sec:experiments:ablation}
    
    \noindent\textbf{Offline and online self-distillation.}
    Table~\ref{tab:searchqa-ablation} isolates the two stages of CAPS. Using only
    offline trajectory self-distillation reaches 37.4 and 42.9 average EM for 3B
    and 7B, while using only online policy self-distillation reaches 37.1 and 41.4.
    Combining them improves the averages to 39.2 and 43.5, showing that the two
    stages provide complementary supervision. We also replace the visual-history
    input in offline distillation with the original text trajectory. This variant
    falls to 36.3 and 40.7 average EM. For 7B, its result is close to AgentOCR
    without distillation (40.1), despite using successful teacher trajectories.
    The comparison shows that offline distillation must initialize the target
    policy with the ability to reason and make decisions from visual history;
    exposure to successful text trajectories alone is not sufficient.\par

    \begin{table}[!htbp]
    \centering
    \caption{{\textbf{SearchQA ablation of our two-stage CAPS.} We report EM (\%) and memory-context token cost per step. 
    \textbf{Off-SD} denotes Offline Trajectory Self-Distillation performed before
    GRPO, whereas \textbf{On-SD} denotes Online Policy Self-Distillation applied
    during GRPO. \textbf{Text-Traj. SFT} uses the original text history instead of
    rendered visual history during offline supervised fine-tuning.
    }}
    \label{tab:searchqa-ablation}
    \small
    \setlength{\tabcolsep}{2pt}
    \renewcommand{\arraystretch}{1.12}
    
    \begin{adjustbox}{max width=\linewidth}
    \begin{tabular}{@{}l*{10}{c}@{}}
    \toprule
    \rowcolor{HeaderGray}
    \textbf{Variant} & \multicolumn{3}{c}{\textbf{Single-Hop}}
    & \multicolumn{4}{c}{\textbf{Multi-Hop}} & \textbf{Overall}
    & \multicolumn{2}{c}{\textbf{Tokens/Step}} \\
    \cmidrule(lr){2-4}\cmidrule(lr){5-8}\cmidrule(lr){9-9}\cmidrule(lr){10-11}
    & NQ & TriviaQA & PopQA & HotpotQA & 2Wiki & MuSiQue
    & Bamboogle & Avg.$\uparrow$ & Avg.$\downarrow$ & Max.$\downarrow$ \\
    \midrule
    \rowcolor{BlockGray}
    \multicolumn{11}{c}{\textit{Qwen2.5-VL-3B-Instruct}} \\
    \textbf{CAPS}
    & \bestvis{42.5} & \bestvis{60.1} & \bestvis{44.4} & \bestvis{40.1}
    & \bestvis{37.8} & \bestvis{15.4} & \bestvis{34.4} & \bestvis{39.2}
    & 0.37k & \textbf{1.76k} \\
    \midrule
    Off-SD $\rightarrow$ GRPO
    & 40.6 & \secondvis{59.5} & 43.2 & \secondvis{39.6} & 36.2
    & \secondvis{14.0} & 28.8 & \secondvis{37.4} & \underline{0.27k} & 3.38k \\
    GRPO + On-SD
    & 41.5 & 56.8 & 43.2 & 36.6 & \secondvis{36.4} & 13.0
    & \secondvis{32.0} & 37.1 & \textbf{0.26k} & \underline{3.32k} \\
    Text-Traj. SFT $\rightarrow$ GRPO
    & \secondvis{42.0} & 59.2 & \secondvis{43.6} & 37.3 & 35.5
    & 11.8 & 24.8 & 36.3 & 0.32k & 3.48k \\
    \midrule
    \rowcolor{BlockGray}
    \multicolumn{11}{c}{\textit{Qwen2.5-VL-7B-Instruct}} \\
    \textbf{CAPS}
    & 44.4 & \secondvis{64.2} & \secondvis{47.1} & \secondvis{42.9}
    & \bestvis{42.3} & \secondvis{18.3} & \bestvis{45.6} & \bestvis{43.5}
    & \underline{0.29k} & 1.74k \\
    \midrule
    Off-SD $\rightarrow$ GRPO
    & \bestvis{45.7} & \bestvis{64.7} & \bestvis{47.8} & \bestvis{43.7}
    & \secondvis{41.1} & \bestvis{18.7} & \secondvis{38.4} & \secondvis{42.9}
    & 0.35k & \underline{1.68k} \\
    GRPO + On-SD
    & 43.7 & 62.5 & 46.3 & 41.8 & 40.0 & 17.8 & 37.6 & 41.4 & 0.32k & 2.26k \\
    Text-Traj. SFT $\rightarrow$ GRPO
    & \secondvis{45.2} & 63.6 & 44.6 & 39.3 & 38.7 & 15.7 & 37.6 & 40.7
    & \textbf{0.21k} & \textbf{0.86k} \\
    \bottomrule
    \end{tabular}
    \end{adjustbox}

    \end{table}

\begin{figure}[!htbp]
    \centering
        \includegraphics[width=0.96\linewidth]{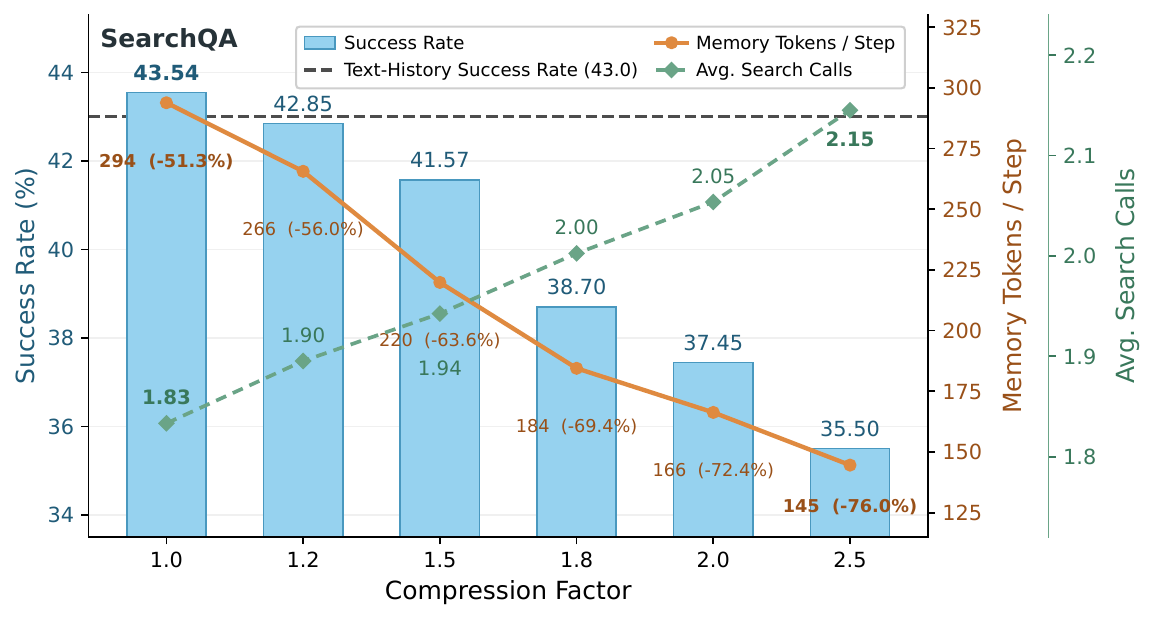}
        
        \caption{\textbf{SearchQA compression trade-off (7B CAPS).} Success and search calls are seven-dataset means. Parentheses show token savings over the text
        teacher; the gray dashed line marks its 43.0\% reference
        from Table~\ref{tab:searchqa-main}.}
        \label{fig:searchqa-compression-tradeoff}
    \par\medskip
    \includegraphics[width=0.96\linewidth]{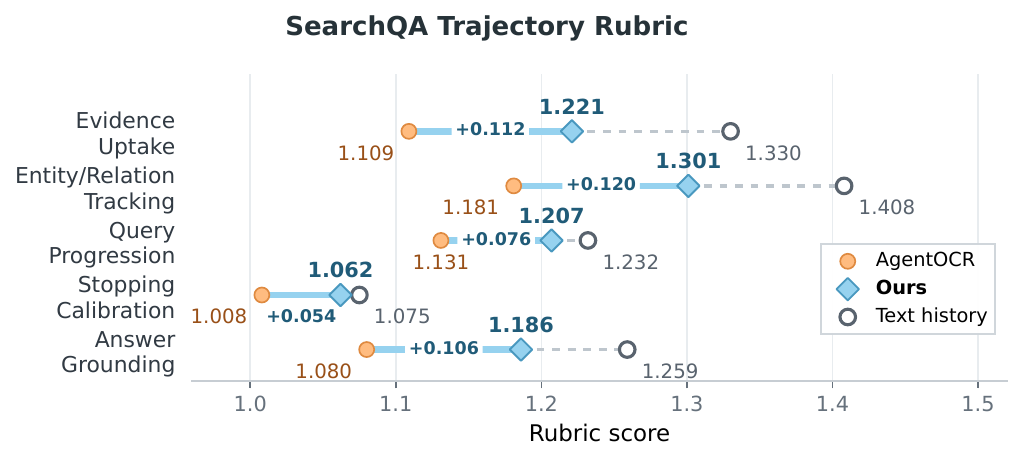}
    \caption{\textbf{Blind trajectory-rubric scores on SearchQA.}
    Scores use inverse-probability weighting (0--2; higher is better).}
    \label{fig:searchqa-rubric}
    
    \end{figure}

    \noindent\textbf{Image compression factor.}
    Following AgentOCR, the compression factor $c\geq 1$ controls additional
    downsampling of the rendered history image. Before minimum-size clipping, an
    image of width $W$ and height $H$ is resized to
    \(W_c=\lfloor W/\sqrt{c}\rfloor\) and
    \(H_c=\lfloor H/\sqrt{c}\rfloor\).
    Its area is therefore reduced by approximately a factor of $c$. We use $c=1$
    by default. As shown in Fig.~\ref{fig:searchqa-compression-tradeoff}, stronger
    compression causes a rapid performance drop: increasing $c$ from 1.0 to 2.5
    reduces SearchQA success from 43.54\% to 35.50\%. Lower resolution makes visual
    history harder to recover, causing the policy to perform more redundant
    searches during reasoning and decision making; the average number of search
    calls increases from 1.83 to 2.15. These additional interactions lengthen the
    history and offset part of the token savings expected from lower image
    resolution. We therefore do not reduce token cost through additional image
    downsampling in the default setting. Instead, CAPS lowers context cost through
    a better decision policy that completes tasks with fewer interactions. Additional ablations on the distillation weight, clipping threshold,
    and divergence are provided in Appendix~\ref{app:additional-ablations}.
    
    \par
    \subsection{Policy-Gap Analysis}
    \label{sec:experiments:policy-gap}
    
    We repeat the diagnostics in Sec.~\ref{sec:diagnosis} to test whether CAPS improves the identified mechanisms.
    
    \noindent\textbf{History recovery.}
    Table~\ref{tab:diagnosis-history-qa} compares generic history recovery with
    end-task performance. CAPS reaches 43.5\% on SearchQA, compared with 40.1\%
    for AgentOCR, but its image-history QA score is lower (73.70 versus 78.22).
    The base VLM obtains the highest image-history QA score of 82.78 while reaching
    only 14.0\% on SearchQA. CAPS therefore does not improve performance by
    strengthening generic OCR or history transcription. Instead, the gain comes
    from using the recovered information more effectively for agent decisions.
    The lower history-QA score may reflect specialization toward the policy task
    rather than general visual question answering.\par

    \begin{table}[!htbp]
    \centering
    \small
    \centering
    \caption{\textbf{History recovery vs. SearchQA Acc.} We report exact match (\%) by question type on the strict History QA benchmark.}
    \label{tab:diagnosis-history-qa}
    \setlength{\tabcolsep}{4pt}
    \renewcommand{\arraystretch}{1.12}
    \begin{adjustbox}{max width=\linewidth}
    \begin{tabular}{lccc}
    \toprule
    \textbf{Metric} & \shortstack{\textbf{Base VLM} \\ \texttt{OCR w/o RL}}
    & \textbf{AgentOCR} & \textbf{CAPS} \\
    \midrule
    SearchQA success & 14.0 & 40.1 & \textbf{43.5} \\
    \midrule
    Document title & \textbf{78.2} & 71.2 & 65.6 \\
    Keyword presence & \textbf{99.3} & 99.2 & 98.3 \\
    Last search query & \textbf{86.4} & 77.6 & 64.3 \\
    Year/number & \textbf{67.2} & 64.9 & 66.6 \\
    Image overall & \textbf{82.78} & 78.22 & 73.70 \\
    Text overall & \textbf{90.03} & 89.78 & 89.35 \\
    \bottomrule
    \end{tabular}
    \end{adjustbox}
    \end{table}

    \begin{table}[!htbp]
    \centering
    \caption{\textbf{Decision alignment under matched semantic states.} Search
    content match is query token-set Jaccard $\geq 0.8$ when both choose Search;
    answer content match is normalized EM when both choose Answer.
    }
    \label{tab:diagnosis-decision-gap}
    \small
    \setlength{\tabcolsep}{4pt}
    \renewcommand{\arraystretch}{1.12}
    \begin{adjustbox}{max width=\linewidth}
    \begin{tabular}{p{0.43\columnwidth}ccc}
    \toprule
    \textbf{Metric (\%)} & \textbf{AgentOCR} & \textbf{CAPS} & \textbf{Gap} \\
    \midrule
    Valid action rate & 99.96 & \textbf{99.97} & +0.01 \\
    Action agreement & 70.87 & \textbf{79.53} & +8.66 \\
    Text Search / Visual Answer & 51.67 & \textbf{32.42} & -19.25 \\
    Text Answer / Visual Search & \textbf{5.37} & 7.85 & +2.48 \\
    Search content match & 5.22 & \textbf{23.53} & +18.31 \\
    Answer content match & 71.82 & \textbf{72.89} & +1.07 \\
    \bottomrule
    \end{tabular}
    \end{adjustbox}
    \end{table}
    
    \noindent\textbf{Matched-state decision alignment.}
    Table~\ref{tab:diagnosis-decision-gap} evaluates the visual-history and
    text-history policies under matched semantic states. CAPS raises action-type
    agreement with the text-history policy from 70.87\% to 79.53\%. More
    importantly, cases in which the text policy searches but the visual policy
    answers fall from 51.67\% to 32.42\%, a reduction of 19.25 points. Search
    content match also increases from 5.22\% to 23.53\%. After distillation, the visual-history policy is less likely to stop before
    collecting the evidence requested by the text-history policy.\par

    \noindent\textbf{Trajectory quality.}
    Figure~\ref{fig:searchqa-rubric} reports blind trajectory ratings
    under the five-dimensional rubric defined in Sec.~\ref{sec:diagnosis}.
    CAPS outperforms AgentOCR on all five dimensions, increasing the
    total score from 5.509 to 5.977 and approaching the text-history
    policy's 6.304.
    Gains in evidence use, query progression, stopping, and answer grounding
    show that CAPS transfers reasoning and decision behavior rather than
    improving visual text recognition.\par

    \section{Conclusion}
    In this paper, we identify an agentic policy gap when multi-step histories are represented as images: visual-history agents can recover content yet use it less effectively for reasoning and decisions. Therefore, we propose CAPS to combine offline trajectory and online policy self-distillation to transfer text-history behavior and improves performance while preserving the token efficiency of vision--text compression. Our results show that effective visual-history compression requires policy alignment, rather than merely preserving recoverable content.

%% file: appendix.tex
\definecolor{HeaderGray}{RGB}{242,245,247}
\definecolor{OursBlue}{RGB}{232,244,250}
\makeatletter
\@fptop=0pt
\@fpsep=12pt
\@fpbot=0pt plus 1fil
\makeatother

\FloatBarrier
\section{Experimental and Implementation Details}
\label{app:data-implementation}

\subsection{Training data construction}

Offline trajectory self-distillation uses successful rollouts produced by a
size-matched text-history policy. These are the actual rollouts saved during
text-history policy training. The text-history and visual-history RL policies
use the same benchmark training split, so every question used to construct the
SFT trajectories already belongs to the task set used by the subsequent visual
policy GRPO stage. Offline distillation therefore adds trajectory supervision
but introduces no additional task instances. For SearchQA, we read trajectories
saved at training step 50 or later. A trajectory is retained only if it succeeds
with episode reward 1, all actions are valid, its normalized step indices are
contiguous, it contains at most four steps, and every response is no longer than
511 tokens. We additionally reject responses with malformed, empty,
conflicting, truncated, or environment-mismatched \texttt{<search>} and
\texttt{<answer>} blocks. The last step must contain a valid final answer.

Several valid rollouts can exist for the same training question. We retain one
trajectory per question by lexicographically preferring fewer interaction
steps, fewer repeated searches, fewer response tokens, fewer prompt tokens,
and finally a later teacher-training step. Every retained trajectory is then
expanded into one SFT example per decision step. The input contains the task
and the rendered prefix before that decision. Specifically, the original
textual interaction history from the text-history teacher trajectory is
rendered into a visual history using the same rendering pipeline as in RL;
the student therefore receives the rendered image history rather than the
original serialized text history. The target is the complete teacher response,
including its reasoning and executable action.

ALFWorld uses the same construction with task-specific validity checks. A
retained trajectory must succeed with reward 10, finish in a won/done state,
contain a contiguous sequence of no more than 30 steps, and have valid
\texttt{<action>} blocks that match the environment actions. When several
trajectories are available for one task instance, we prefer fewer steps,
shorter responses, shorter prompts, and a later training step. Both benchmarks
use a deterministic 98/2 split by task identifier, so decision steps from one
task never cross the SFT train and validation splits.

\begin{table}[!hbp]
\centering
\caption{Offline distillation data after filtering and one-best-trajectory
selection. ``Samples'' counts step-level SFT examples.}
\label{tab:app-offline-data}
\small
\renewcommand{\arraystretch}{1.12}
\begin{adjustbox}{max width=\linewidth}
\begin{tabular}{llrrrr}
\toprule
\textbf{Benchmark} & \textbf{Model size} & \textbf{Selected tasks}
& \textbf{Samples} & \textbf{Train} & \textbf{Validation} \\
\midrule
SearchQA & 3B & 7,734 & 18,005 & 17,599 & 406 \\
SearchQA & 7B & 7,899 & 18,920 & 18,487 & 433 \\
ALFWorld & 3B & 1,499 & 15,011 & 14,780 & 231 \\
ALFWorld & 7B & 1,460 & 15,005 & 14,679 & 326 \\
\bottomrule
\end{tabular}
\end{adjustbox}
\end{table}
Table~\ref{tab:app-offline-data} summarizes the resulting filtered datasets.
SearchQA data contain only NQ~\citep{kwiatkowski2019natural} and
HotpotQA~\citep{yang2018hotpotqa} training questions. The 3B data contain
3,215 NQ and 4,519 HotpotQA identifiers, while the 7B data contain 3,299 and
4,600, respectively. The remaining five SearchQA datasets are not used to
construct offline distillation examples. For ALFWorld, the data cover all six
task families.

\subsection{Rendering, optimization, and runtime configuration}

For both benchmarks, we follow the benchmark and rendering configurations
released with AgentOCR~\citep{feng2026agentocr}. Its
ALFWorld~\citep{shridhar2020alfworld} setup follows the default GiGPO
configuration~\citep{feng2026group}: optical and text prompts are
limited to 2,048 and 5,120 tokens, respectively; responses are limited to 512
tokens; each update uses 16 questions with eight rollouts per question; the
episode horizon is 50 steps with full history; successful episodes receive
reward 10; rollout and validation temperatures are 1.0 and 0.4. Following
AgentOCR and SKILL0~\citep{feng2026agentocr,lu2026skill0}, our main ALFWorld
results use $H=50$, and we additionally evaluate $H=2$, the GiGPO default
history window.

The SearchQA setup follows AgentOCR's Search-R1-style
configuration~\citep{feng2026agentocr,jin2025search}. Optical and text prompts
are limited to 4,096 and 14,000 tokens, respectively; responses are limited to
512 tokens; each update uses 128 questions with eight rollouts per question;
the episode horizon is four steps with full history; correct answers receive
reward 1; rollout temperature is 1.0 and validation is greedy. Both benchmarks
use actor learning rate $10^{-6}$, PPO mini-batch size 256, and no
reference-policy KL penalty in the GRPO objective~\citep{shao2024deepseekmath}.

Rendering also follows AgentOCR~\citep{feng2026agentocr}. Text uses a monospace
font and line spacing 1.2. SearchQA uses font size 12, maximum width 560 pixels,
and five pixels of final padding; Search tags are blue and Information tags are
red. ALFWorld uses font size 10, maximum width 392 pixels, and the same padding;
Observations are blue and Actions are red. Step zero receives the same minimally
padded blank image used by the runtime pipeline.

Runtime caching follows AgentOCR~\citep{feng2026agentocr}. Unchanged history
content is divided into reusable rendered units, cached once, and assembled in
chronological order for the current policy input. This avoids repeatedly
rendering identical earlier history without changing the information presented
to the model.\looseness=-1\par

\begin{table}[!htbp]
\centering
\captionsetup{skip=6pt}
\caption{Training hyperparameters for offline trajectory self-distillation,
visual-history GRPO, and online policy self-distillation. Values are shared by
3B and 7B unless a row explicitly reports them as 3B/7B.}
\label{tab:app-hyperparameters}
\small
\renewcommand{\arraystretch}{1.12}
\begin{adjustbox}{max width=\linewidth}
\begin{tabular}{@{}p{0.43\textwidth}*{2}{>{\centering\arraybackslash}p{0.25\textwidth}}@{}}
\toprule
\textbf{Parameter} & \textbf{SearchQA} & \textbf{ALFWorld} \\
\midrule
\rowcolor{HeaderGray}\multicolumn{3}{l}{\textbf{Offline trajectory self-distillation}} \\
Global / per-GPU micro batch & 32 / 1 & 16 / 1 \\
Maximum sequence length & 8,192 & 2,560 \\
LoRA rank / alpha / target & 32 / 64 / all-linear & 32 / 64 / all-linear \\
Optimizer / $\beta_1,\beta_2$ & AdamW / 0.9, 0.95 & AdamW / 0.9, 0.95 \\
Learning rate / warmup / weight decay & $2\!\times\!10^{-4}$ / 0.03 / 0 & $2\!\times\!10^{-4}$ / 0.03 / 0 \\
Scheduler / gradient clipping / precision & cosine / 1.0 / bf16 & cosine / 1.0 / bf16 \\
Configured training budget & 2 epochs & 2 epochs \\
Training GPUs (3B / 7B) & 2 / 4 & 4 / 4 \\
\midrule
\rowcolor{HeaderGray}\multicolumn{3}{l}{\textbf{Visual-history GRPO}} \\
Questions/update / rollouts/question & 128 / 8 & 16 / 8 \\
Maximum environment steps & 4 & 50 \\
Prompt limits (image / teacher text) & 4,096 / 14,000 & 2,048 / 5,120 \\
Maximum response length & 512 & 512 \\
Actor learning rate / PPO epochs & $10^{-6}$ / 1 & $10^{-6}$ / 1 \\
PPO mini / per-GPU micro batch & 256 / 8 & 256 / 8 \\
PPO clip / dual clip / entropy coefficient & 0.2 / 3.0 / 0.001 & 0.2 / 3.0 / 0.001 \\
Success / failure reward & 1 / 0 & 10 / 0 \\
Reference-policy KL coefficient & 0 & 0 \\
Rollout / validation temperature & 1.0 / 0.0 & 1.0 / 0.4 \\
Maximum optimization steps & 150 & 150 \\
Training GPUs (3B / 7B) & 2 / 4 & 4 / 4 \\
\midrule
\rowcolor{HeaderGray}\multicolumn{3}{l}{\textbf{Online policy self-distillation}} \\
Teacher temperature & 1.1 & 1.1 \\
Divergence / vocabulary support & forward KL / full & forward KL / full \\
Distillation weight $\lambda$ / per-token clip $\tau$ & 0.05 / 0.05 & 0.01 / 0.05 \\
Rollout filter / supervised span & valid actions / full response & valid actions / full response \\
Teacher precision & bf16 & bf16 \\
\bottomrule
\end{tabular}
\end{adjustbox}
\end{table}

All unqualified values in Table~\ref{tab:app-hyperparameters} are shared by 3B
and 7B. Scale-dependent settings are reported explicitly in 3B/7B order and
are limited to GPU allocation. ``Rollout filter'' means that online
distillation is applied only when the student response contains a valid parsed
action; invalid-action responses receive no distillation loss. ``Supervised
span'' means that the divergence is computed over the complete generated
response, including both reasoning tokens and the executable action, rather
than only the action span.

LoRA targets all matching linear layers, including matching layers in the
visual stack; the vision encoder is therefore not frozen. In the notation of
the main-paper joint objective, $\lambda$ is the coefficient of the online
distillation loss, while $\tau$ is the per-token KL clipping threshold in the
online loss. The online teacher is frozen, evaluated in bf16, and scores the
complete student-generated response under the corresponding text-history
prompt. It is removed for validation and inference.

Experiments ran on NVIDIA H200 GPUs. SearchQA uses two GPUs for 3B and four for
7B RL, while ALFWorld uses four GPUs for both scales. All two-epoch SFT runs
complete within three hours. For either benchmark, a complete RL
run for one model scale typically takes approximately one to two days.
CAPS incurs additional training cost from offline trajectory
self-distillation and online teacher supervision. At inference,
only the visual-history student is used, with no additional
distillation-specific computation. With identical inputs and generation settings, AgentOCR and CAPS
require 74.20 and 52.06 minutes of cumulative generation time,
respectively, for 10,000 single-step decisions. The reduction is
consistent with the shorter responses generated by CAPS.

\FloatBarrier
\section{Additional Experiments}
\label{app:additional-ablations}

\subsection{SearchQA Parameter Ablations}

We report additional experiments on the distillation weight, clipping
threshold, and divergence on SearchQA with Qwen2.5-VL-3B. The initialization checkpoints,
training budget, and evaluation protocol follow the experimental configuration
described above.

The main method uses forward KL, $\lambda=0.05$, and $\tau=0.05$.
We examine $\lambda\in\{0.01,0.10\}$, $\tau\in\{0.01,0.10\}$, removal of
the distillation cap, and alternative divergences. Removing the cap leaves
GRPO clipping unchanged. The divergence variants use reverse KL and symmetric
JSD, retaining $\lambda=0.05$ and $\tau=0.05$.

Following the divergence formulation used in on-policy
distillation~\citep{zhao2026self}, we use the student and teacher distributions
$p_{b,k}^{S}$ and $p_{b,k}^{T}$ from
Eq.~\eqref{eq:method:online-distributions}, with the sample index $b$
made explicit as in Eq.~\eqref{eq:method:online-kl}. Reverse KL is
\begin{equation}
D_{b,k}^{\mathrm{RKL}}
=\sum_{v\in\mathcal{V}}p_{b,k}^{S}(v)
\log\frac{p_{b,k}^{S}(v)}{p_{b,k}^{T}(v)}.
\label{eq:app:reverse-kl}
\end{equation}
For symmetric JSD, let $q_{b,k}(v)=\tfrac12
\bigl(p_{b,k}^{S}(v)+p_{b,k}^{T}(v)\bigr)$. Then
\begin{equation}
D_{b,k}^{\mathrm{JSD}}
=\frac12\sum_{v\in\mathcal{V}}
\left[
p_{b,k}^{S}(v)\log\frac{p_{b,k}^{S}(v)}{q_{b,k}(v)}
+p_{b,k}^{T}(v)\log\frac{p_{b,k}^{T}(v)}{q_{b,k}(v)}
\right].
\label{eq:app:symmetric-jsd}
\end{equation}
Each variant replaces $D_{b,k}^{\mathrm{FKL}}$ in
Eq.~\eqref{eq:method:online-loss}, with the same response mask and clipping
threshold.

\begin{table}[!htbp]
\centering
\small
\caption{Additional SearchQA 3B results. Bold and underline mark the best
and second-best accuracy in each column. AgentOCR results
follow the main table~\citep{feng2026agentocr}.}
\label{tab:app-parameter-ablations}
\begin{adjustbox}{max width=\linewidth}
\begin{tabular}{@{}l*{9}{r}@{}}
\toprule
Setting & NQ & TriviaQA & PopQA & HotpotQA & 2Wiki & MuSiQue & Bamboogle
& Avg.$\uparrow$ & \shortstack{History tokens\\per step$\downarrow$} \\
\midrule
AgentOCR & 38.6 & 56.5 & 41.7 & 33.6 & 30.7 & 14.6 & 24.0 & 34.2 & 260 \\
\rowcolor{OursBlue}CAPS & 42.5 & 60.1 & \underline{44.4} & \underline{40.1} & 37.8 & \textbf{15.4} & \textbf{34.4} & \textbf{39.24} & 369.77 \\
\midrule
$\lambda=0.01$ & \textbf{43.7} & 60.0 & 44.0 & 39.6 & 34.5 & 12.6 & 26.4 & 37.24 & 347.45 \\
$\lambda=0.10$ & \underline{43.2} & \underline{60.5} & \underline{44.4} & \textbf{40.3} & 35.9 & \underline{15.0} & 30.4 & \underline{38.52} & 439.02 \\
\midrule
$\tau=0.01$ & 42.8 & 59.7 & 43.7 & 38.9 & 36.0 & 13.3 & \underline{32.8} & 38.18 & 323.99 \\
$\tau=0.10$ & 42.0 & 60.3 & 44.2 & 39.9 & \underline{37.9} & 14.7 & 28.0 & 38.15 & 334.01 \\
No clipping & 42.1 & \textbf{60.9} & \textbf{45.2} & 39.7 & \textbf{38.8} & 13.4 & 25.6 & 37.99 & 330.21 \\
\midrule
Reverse KL & 42.2 & 59.6 & 44.1 & 39.7 & 36.2 & 14.9 & 32.0 & 38.39 & 331.13 \\
JSD & 42.7 & 59.6 & 40.6 & 39.1 & 36.2 & 14.2 & 30.4 & 37.56 & 342.82 \\
\bottomrule
\end{tabular}
\end{adjustbox}
\end{table}

Table~\ref{tab:app-parameter-ablations} reports the main result alongside the
additional runs and the reported AgentOCR baseline. The variants reach
37.24--38.52\% average accuracy, exceeding AgentOCR's reported 34.2\% while
remaining below CAPS.\looseness=-1\par

\noindent\textbf{Distillation weight.}
The $\lambda=0.10$ run achieves the highest average among the additional
variants, but remains 0.72 points below CAPS. The $\lambda=0.01$ run is
2.00 points lower. Neither increasing nor decreasing the distillation weight
improves on the main result in these runs.

\noindent\textbf{Clipping and divergence.}
The altered clipping settings have accuracy gaps of 1.06--1.25 points
relative to CAPS. Reverse KL and JSD trail the main result by 0.85 and
1.68 points, respectively. None of these alternatives exceeds the main
forward-KL configuration.

\FloatBarrier
\subsection{Additional Model Results}
\label{app:additional-models}

We additionally evaluate CAPS with Qwen3-VL-4B-Instruct using the same
two-stage pipeline of offline trajectory self-distillation and online policy
self-distillation. Table~\ref{tab:app-additional-models} compares CAPS with
AgentOCR across models and reports history-context costs. With Qwen3-VL-4B, CAPS
achieves 40.4\% average EM on SearchQA with 0.27k history tokens per step,
and 96.5\% success on full-history ALFWorld with 0.18k history tokens per step.
CAPS achieves higher overall scores than the corresponding AgentOCR
baselines on both benchmarks.

\begin{table}[!htbp]
\centering
\caption{AgentOCR and CAPS across models. $^{\dagger}$AgentOCR results from
\citet{feng2026agentocr}.}
\label{tab:app-additional-models}
\small
\renewcommand{\arraystretch}{1.12}
\setlength{\tabcolsep}{2pt}
\begin{adjustbox}{max width=\linewidth}
\begin{tabular}{@{}l*{10}{c}@{}}
\toprule
\rowcolor{HeaderGray}\multicolumn{11}{l}{\rule{0pt}{2.4ex}\textbf{SearchQA}} \\
\textbf{Method}
& \multicolumn{3}{c}{\textbf{Single-Hop}}
& \multicolumn{4}{c}{\textbf{Multi-Hop}}
& \textbf{Overall}
& \multicolumn{2}{c}{\textbf{Tokens/Step}} \\
\cmidrule(lr){2-4}
\cmidrule(lr){5-8}
\cmidrule(lr){9-9}
\cmidrule(lr){10-11}
& NQ & TriviaQA & PopQA & HotpotQA & 2Wiki & MuSiQue & Bamboogle
& Avg.$\uparrow$ & Avg.$\downarrow$ & Max.$\downarrow$ \\
\midrule
\multicolumn{11}{c}{\textit{Qwen2.5-VL-3B-Instruct}} \\
AgentOCR$^{\dagger}$ & 38.6 & 56.5 & 41.7 & 33.6 & 30.7 & 14.6 & 24.0
& 34.2 & 0.26k & 2.50k \\
CAPS & 42.5 & 60.1 & 44.4 & 40.1 & 37.8 & 15.4 & 34.4
& 39.2 & 0.37k & 1.76k \\
\midrule
\multicolumn{11}{c}{\textit{Qwen2.5-VL-7B-Instruct}} \\
AgentOCR$^{\dagger}$ & 43.1 & 61.0 & 45.4 & 40.8 & 38.3 & 15.7 & 36.8
& 40.1 & 0.36k & 2.65k \\
CAPS & 44.4 & 64.2 & 47.1 & 42.9 & 42.3 & 18.3 & 45.6
& 43.5 & 0.29k & 1.74k \\
\midrule
\multicolumn{11}{c}{\textit{Qwen3-VL-4B-Instruct}} \\
AgentOCR & 35.4 & 58.0 & 40.8 & 34.8 & 33.2 & 10.5 & 26.4
& 34.1 & 0.25k & 3.28k \\
CAPS & 41.3 & 62.2 & 45.2 & 40.6 & 38.2 & 15.6 & 40.0
& 40.4 & 0.27k & 1.12k \\
\midrule
\rowcolor{HeaderGray}\multicolumn{11}{l}{\rule{0pt}{2.4ex}\textbf{ALFWorld}} \\
\textbf{Method} & \textbf{$H$}
& \textbf{Pick} & \textbf{Look} & \textbf{Clean}
& \textbf{Heat} & \textbf{Cool} & \textbf{Pick2}
& \textbf{Avg.$\uparrow$}
& \multicolumn{2}{c}{\textbf{Tokens/Step}} \\
\cmidrule(lr){10-11}
& & & & & & & & & Avg.$\downarrow$ & Max.$\downarrow$ \\
\midrule
\multicolumn{11}{c}{\textit{Qwen2.5-VL-3B-Instruct}} \\
AgentOCR$^{\dagger}$ & 50 & 91.9 & 81.8 & 76.0 & 73.3 & 76.1 & 70.0
& 78.2 & 0.38k & 1.14k \\
CAPS & 50 & 94.4 & 91.7 & 97.2 & 100.0 & 93.0 & 83.8
& 93.8 & 0.20k & 0.95k \\
\midrule
\multicolumn{11}{c}{\textit{Qwen2.5-VL-7B-Instruct}} \\
AgentOCR$^{\dagger}$ & 50 & 95.6 & 96.2 & 78.1 & 73.2 & 72.4 & 72.0
& 81.2 & 0.43k & 1.22k \\
CAPS & 50 & 96.9 & 96.7 & 100.0 & 92.3 & 100.0 & 87.7
& 95.7 & 0.21k & 0.87k \\
\midrule
\multicolumn{11}{c}{\textit{Qwen3-VL-4B-Instruct}} \\
AgentOCR & 50 & 95.3 & 65.8 & 87.9 & 58.3 & 63.5 & 56.7
& 74.6 & 0.20k & 0.69k \\
CAPS & 50 & 100.0 & 100.0 & 100.0 & 85.8 & 94.1 & 96.7
& 96.5 & 0.18k & 0.64k \\
\bottomrule
\end{tabular}
\end{adjustbox}
\end{table}

\FloatBarrier
\section{Diagnostic Evaluation Details and Full Results}
\label{app:diagnostics}

\subsection{History recovery}

The strict History QA set is built from real trajectories saved during
SearchQA training. For every noninitial decision, we reconstruct all prior
interaction history that was visible to the model when making that decision.
Because SearchQA permits at most four decision steps, these examples contain
the Search actions and returned results from the preceding one, two, or three
steps. We retain histories containing a search action and returned information,
and generate four extraction question types: document title, keyword presence,
most recent search query, and year/number. Questions whose task identifier
appeared in the selected offline SFT data are excluded from the strict split.

The construction scanned 64,578 trajectories. After excluding task identifiers
used for offline trajectory self-distillation and removing decisions without a
non-empty Search--Information history, balanced sampling produced 4,000
questions from 2,104 task identifiers,
with exactly 1,000 questions per type. The set contains 2,258 HotpotQA and
1,742 NQ examples; 1,268 examples come from successful trajectories and 3,768
from trajectories whose actions were all valid. The step distribution is
2,860/959/181 for steps 1/2/3. Image and text modes use the same question and
history content, changing only whether the history is rendered.

\begin{table}[!htbp]
\centering
\caption{Overall results on the strict History QA benchmark. The EM gap is
Text EM minus Image EM. Base VLM denotes \texttt{OCR w/o RL}.}
\label{tab:app-history-overall}
\small
\begin{adjustbox}{max width=\linewidth}
\begin{tabular}{lrrr}
\toprule
\textbf{Metric} & \textbf{Base VLM} & \textbf{AgentOCR} & \cellcolor{OursBlue} \textbf{\capsmodel} \\
\midrule
Image EM & 82.78 & 78.22 & \cellcolor{OursBlue}73.70 \\
Image F1 & 87.30 & 84.77 & \cellcolor{OursBlue}83.49 \\
Text EM & 90.03 & 89.78 & \cellcolor{OursBlue}89.35 \\
Text F1 & 91.34 & 91.24 & \cellcolor{OursBlue}90.65 \\
Text--Image EM gap & 7.25 & 11.56 & \cellcolor{OursBlue}15.65 \\
\bottomrule
\end{tabular}
\end{adjustbox}
\end{table}

\begin{table}[!htbp]
\centering
\caption{History QA EM by source dataset. Base VLM denotes
\texttt{OCR w/o RL}.}
\label{tab:app-history-source}
\small
\begin{adjustbox}{max width=\linewidth}
\begin{tabular}{lrrr}
\toprule
\textbf{Source Dataset} & \textbf{Base VLM} & \textbf{AgentOCR} & \cellcolor{OursBlue} \textbf{\capsmodel} \\
\midrule
\multicolumn{4}{l}{\textbf{Image History}} \\
HotpotQA & 79.19 & 74.40 & \cellcolor{OursBlue}71.04 \\
NQ & 87.43 & 83.18 & \cellcolor{OursBlue}77.15 \\
\midrule
\multicolumn{4}{l}{\textbf{Text History}} \\
HotpotQA & 87.47 & 87.29 & \cellcolor{OursBlue}86.40 \\
NQ & 93.34 & 93.00 & \cellcolor{OursBlue}93.17 \\
\bottomrule
\end{tabular}
\end{adjustbox}
\end{table}

``Image'' evaluates each model from the rendered history, whereas ``Text''
provides the same interaction history directly as text, without an image input.
Although \capsmodel improves downstream SearchQA, it does not improve generic
history extraction, so stronger OCR cannot explain the task gains.

\subsection{Matched-state decisions}

We extract all nonempty decision prefixes from the text-history validation
trajectories. From 51,713 trajectories, this yields 99,969 prefixes; six with
invalid text actions are removed. We then draw a 10,000-prefix sample
stratified by dataset, step, text action type, and text-trajectory success.
The sample contains 5,123 text Search states and 4,877 text Answer states. The
two visual policies receive the same question, canonical history content,
rendered image, and native SearchQA prompt. Both produce all 10,000 outputs
with no missing, duplicate, extra, or input-mismatched prefixes. Generation is
deterministic with temperature 0, top-$p$ 1, top-$k=-1$, and at most 512
response tokens.\looseness=-1\par

Valid-action rate measures the fraction of visual outputs parsed as Search or
Answer. Action-type agreement measures agreement with the text policy, and
the cross-action rates $P(\text{Visual Answer}\mid\text{Text Search})$ and
$P(\text{Visual Search}\mid\text{Text Answer})$ quantify directional disagreement.

For states where both policies Search, we report query exact match and
token-set Jaccard after lowercasing and alphanumeric tokenization.
Search joint alignment counts states where both policies Search and query
Jaccard is at least 0.8, divided by all text-Search states. Answer EM measures
normalized exact match between the visual and text-policy answers when both
policies Answer. Answer joint alignment divides the same matching count by
all text-Answer states.

Hard alignment requires matching action types and either query Jaccard
$\geq0.8$ for Search or answer EM for Answer. Soft alignment averages the
corresponding Jaccard or EM scores over all states; mismatched or invalid
actions contribute zero. Gap denotes \capsmodel{} minus AgentOCR, expressed
in percentage points for rates. In the breakdowns, $N$ counts sampled
prefixes, answer rate is the frequency of Answer actions, and agreement gain
is the difference in action-type agreement between \capsmodel{} and AgentOCR.\looseness=-1\par

\begin{table}[!htbp]
\centering
\caption{Complete matched-state action and content alignment.}
\label{tab:app-decision-core}
\small
\renewcommand{\arraystretch}{1.12}
\begin{adjustbox}{max width=\linewidth}
\begin{tabular}{p{0.45\columnwidth}rrr}
\toprule
\textbf{Metric (\%)} & \textbf{AgentOCR} & \cellcolor{OursBlue}\textbf{\capsmodel} & \textbf{Gap} \\
\midrule
Valid action & 99.96 & \cellcolor{OursBlue}\textbf{99.97} & +0.01 \\
Action-type agreement & 70.87 & \cellcolor{OursBlue}\textbf{79.53} & +8.66 \\
Text Search / Visual Answer & 51.67 & \cellcolor{OursBlue}\textbf{32.42} & -19.25 \\
Text Answer / Visual Search & \textbf{5.37} & \cellcolor{OursBlue}7.85 & +2.48 \\
Query exact match $\mid$ both Search & 1.82 & \cellcolor{OursBlue}\textbf{4.88} & +3.06 \\
Query Jaccard $\geq0.8\mid$ both Search & 5.22 & \cellcolor{OursBlue}\textbf{23.53} & +18.31 \\
Search joint alignment & 2.52 & \cellcolor{OursBlue}\textbf{15.89} & +13.37 \\
Answer EM $\mid$ both Answer & 71.82 & \cellcolor{OursBlue}\textbf{72.89} & +1.07 \\
Answer joint alignment & \textbf{67.95} & \cellcolor{OursBlue}67.15 & -0.80 \\
Hard action-content alignment & 34.43 & \cellcolor{OursBlue}\textbf{40.89} & +6.46 \\
Soft action-content alignment & 42.66 & \cellcolor{OursBlue}\textbf{50.65} & +7.99 \\
\bottomrule
\end{tabular}
\end{adjustbox}
\end{table}

\begin{table}[!htbp]
\centering
\caption{Answer rates and action-type agreement gain by interaction step.}
\label{tab:app-decision-step}
\small
\renewcommand{\arraystretch}{1.12}
\begin{adjustbox}{max width=\linewidth}
\begin{tabular}{rrrrrr}
\toprule
\textbf{Step} & \textbf{$N$} & \shortstack{\textbf{Text}\\\textbf{answer rate}} & \shortstack{\textbf{AgentOCR}\\\textbf{answer rate}} & \shortstack{\textbf{CAPS}\\\textbf{answer rate}}
& \shortstack{\textbf{Agreement}\\\textbf{gain}} \\
\midrule
1 & 5,172 & 29.80 & 55.41 & 39.98 & +11.60 \\
2 & 3,627 & 66.94 & 89.85 & 81.42 & +6.26 \\
3 & 1,201 & 75.60 & 94.59 & 94.34 & +3.25 \\
\bottomrule
\end{tabular}
\end{adjustbox}
\end{table}

\begin{table}[!htbp]
\centering
\caption{Answer rates and action-type agreement by SearchQA dataset.}
\label{tab:app-decision-dataset}
\small
\renewcommand{\arraystretch}{1.12}
\begin{adjustbox}{max width=\linewidth}
\begin{tabular}{lrrrrrr}
\toprule
\textbf{Dataset} & \textbf{$N$} & \shortstack{\textbf{Text}\\\textbf{answer rate}} & \shortstack{\textbf{AgentOCR}\\\textbf{answer rate}}
& \shortstack{\textbf{CAPS}\\\textbf{answer rate}} & \shortstack{\textbf{AgentOCR}\\\textbf{agreement}} & \shortstack{\textbf{CAPS}\\\textbf{agreement}} \\
\midrule
2WikiMultihopQA & 3,141 & 32.79 & 63.42 & 54.22 & 64.63 & 75.77 \\
Bamboogle & 25 & 44.00 & 56.00 & 56.00 & 80.00 & 88.00 \\
HotpotQA & 1,541 & 46.98 & 68.33 & 54.25 & 73.13 & 82.80 \\
MuSiQue & 610 & 34.26 & 58.36 & 39.34 & 71.48 & 87.38 \\
NQ & 531 & 67.98 & 81.17 & 79.10 & 76.27 & 77.40 \\
PopQA & 2,387 & 59.20 & 82.20 & 70.26 & 72.22 & 79.64 \\
TriviaQA & 1,765 & 63.97 & 82.32 & 71.61 & 76.20 & 81.02 \\
\bottomrule
\end{tabular}
\end{adjustbox}
\end{table}

Repeat rates are computed over each visual policy's Search actions. An exact
repeat matches a previous normalized query; a fuzzy repeat has token-set
Jaccard at least 0.8. Response statistics cover all 10,000 outputs and count the
complete response, including reasoning and the executable action.

\begin{table}[!htbp]
\centering
\caption{Repeated search and response-length statistics on matched states.}
\label{tab:app-decision-cost}
\small
\renewcommand{\arraystretch}{1.12}
\begin{adjustbox}{max width=\linewidth}
\begin{tabular}{lrrr}
\toprule
\textbf{Metric} & \textbf{AgentOCR} & \cellcolor{OursBlue}\textbf{\capsmodel} & \textbf{Gap} \\
\midrule
Exact repeated query (\%) & 5.70 & \cellcolor{OursBlue}\textbf{0.31} & -5.39 \\
Fuzzy repeated query (\%) & 10.16 & \cellcolor{OursBlue}\textbf{3.83} & -6.33 \\
Mean response tokens & 167.66 & \cellcolor{OursBlue}\textbf{68.70} & -98.96 \\
Median response tokens & 159 & \cellcolor{OursBlue}\textbf{64} & -95 \\
95th-percentile response tokens & 302 & \cellcolor{OursBlue}\textbf{119} & -183 \\
Hit 512-token limit (\%) & 0.05 & \cellcolor{OursBlue}\textbf{0.03} & -0.02 \\
\bottomrule
\end{tabular}
\end{adjustbox}
\end{table}

On matched histories, \capsmodel{} increases action-type agreement with the
text policy from 70.87\% to 79.53\%
(Table~\ref{tab:app-decision-core}). When the text policy chooses Search,
the visual Answer rate falls from 51.67\% to 32.42\%, indicating closer
alignment in the decision to continue retrieval. Agreement gains are positive
across all three evaluated steps and all seven datasets, with the largest
step-wise gain at the first decision after retrieval (+11.60 points;
Tables~\ref{tab:app-decision-step} and~\ref{tab:app-decision-dataset}).

Retrieval content also becomes more aligned: among states where both policies
Search, the fraction with query Jaccard $\geq0.8$ rises from 5.22\% to
23.53\%. Hard and soft action-content alignment improve by 6.46 and 7.99
points, respectively, supporting closer alignment in both action choice and
retrieval content. Exact and fuzzy repeated-query rates fall from 5.70\%
to 0.31\% and from 10.16\% to 3.83\%
(Table~\ref{tab:app-decision-cost}), indicating less repetitive retrieval.

\subsection{Trajectory rubric}

The trajectory evaluation uses 2,800 questions sampled from 162 strata defined
by correctness pattern, dataset, and cross-policy tool-call differences. Rare
strata are oversampled. Weighted aggregate statistics use inverse-probability
weights whose total corresponds to the 50,505 aligned-question population;
the error-tag table reports raw counts.

For each question, the complete trajectories are assigned balanced anonymous
labels. The judge does not receive policy names, modality labels, gold answers,
or success indicators. It sees the canonical retrieval text and scores
evidence uptake, entity/relation tracking, query progression, stopping
calibration, and answer grounding from 0 to 2. The primary judge is
DeepSeek-V4-Flash~\citep{deepseekai2026deepseekv4} with thinking enabled, temperature 0,
seed 42, and balanced
label order. The immutable finalization contains 2,795 valid core judgments
(99.82\% coverage) and 2,667 judgments with valid step-level metadata
(95.25\% coverage). Five provider content-filter failures are excluded. The
tables report the three policies used in the main diagnostic comparison.

Total sums all five dimensions and ranges from 0 to 10. Local Semantic sums
Evidence, Entity, and Grounding and ranges from 0 to 6; Global Control sums
Query and Stop and ranges from 0 to 4.

\begin{table}[!htbp]
\centering
\caption{Complete primary trajectory-rubric results. Each base dimension
ranges from 0 to 2.}
\label{tab:app-rubric-full}
\small
\begin{adjustbox}{max width=\linewidth}
\begin{tabular}{lrrr}
\toprule
\textbf{Metric} & \textbf{AgentOCR} & \cellcolor{OursBlue}\textbf{\capsmodel} & \textbf{Text History} \\
\midrule
Evidence & 1.109 & \cellcolor{OursBlue}1.221 & \textbf{1.330} \\
Entity & 1.181 & \cellcolor{OursBlue}1.301 & \textbf{1.408} \\
Query & 1.131 & \cellcolor{OursBlue}1.207 & \textbf{1.232} \\
Stop & 1.008 & \cellcolor{OursBlue}1.062 & \textbf{1.075} \\
Grounding & 1.080 & \cellcolor{OursBlue}1.186 & \textbf{1.259} \\
Total & 5.509 & \cellcolor{OursBlue}5.977 & \textbf{6.304} \\
Local Semantic & 3.370 & \cellcolor{OursBlue}3.708 & \textbf{3.997} \\
Global Control & 2.139 & \cellcolor{OursBlue}2.269 & \textbf{2.307} \\
\bottomrule
\end{tabular}
\end{adjustbox}
\end{table}

\begin{table}[!htbp]
\centering
\caption{Deterministic statistics on the same weighted 2,800-question sample.
Success is weighted trajectory success (\%);
Earlier-query matches use Jaccard $\geq0.8$. Entity reuse measures follow-up searches
that reuse a retrieved entity.}
\label{tab:app-rubric-deterministic}
\small
\renewcommand{\arraystretch}{1.12}
\begin{adjustbox}{max width=\linewidth}
\begin{tabular}{lrrr}
\toprule
\textbf{Metric} & \textbf{AgentOCR} & \cellcolor{OursBlue} \textbf{\capsmodel} & \textbf{Text history} \\
\midrule
Trajectory success (\%) & 44.71 & \cellcolor{OursBlue}47.38 & 47.83 \\
Search calls per trajectory & 1.554 & \cellcolor{OursBlue}1.720 & 1.920 \\
Steps per trajectory & 2.554 & \cellcolor{OursBlue}2.720 & 2.920 \\
Earlier-query matches ($\times100$) & 9.80 & \cellcolor{OursBlue}\textbf{3.94} & 7.39 \\
Prompt tokens per step & 427.3 & \cellcolor{OursBlue}453.8 & 703.7 \\
Response tokens per step & 161.4 & \cellcolor{OursBlue}\textbf{67.8} & 68.2 \\
Total tokens per trajectory & 1,576.32 & \cellcolor{OursBlue}\textbf{1,492.74} & 2,407.65 \\
Entity reuse in follow-up searches (\%) & 28.53 & \cellcolor{OursBlue}37.12 & 43.05 \\
\bottomrule
\end{tabular}
\end{adjustbox}
\end{table}

\begin{table}[!htbp]
\centering
\caption{Primary-judge error-tag counts.}
\label{tab:app-rubric-errors}
\small
\renewcommand{\arraystretch}{1.12}
\begin{adjustbox}{max width=\linewidth}
\begin{tabular}{lrrr}
\toprule
\textbf{Error tag} & \textbf{AgentOCR} & \cellcolor{OursBlue}\textbf{CAPS} & \textbf{Text} \\
\midrule
No relevant evidence retrieved & 191 & \cellcolor{OursBlue}\textbf{120} & 156 \\
Evidence present but not used & 545 & \cellcolor{OursBlue}500 & \textbf{432} \\
Entity drift & 913 & \cellcolor{OursBlue}764 & \textbf{611} \\
Relation composition error & 55 & \cellcolor{OursBlue}58 & \textbf{52} \\
Redundant confirmation & \textbf{433} & \cellcolor{OursBlue}560 & 832 \\
Premature answer & 946 & \cellcolor{OursBlue}823 & \textbf{645} \\
Late stop & 25 & \cellcolor{OursBlue}\textbf{19} & 60 \\
Answer extraction error & 722 & \cellcolor{OursBlue}723 & \textbf{613} \\
Contradicts evidence & 401 & \cellcolor{OursBlue}335 & \textbf{216} \\
Invalid action & 3 & \cellcolor{OursBlue}\textbf{2} & \textbf{2} \\
\bottomrule
\end{tabular}
\end{adjustbox}
\end{table}

The rubric results place CAPS between AgentOCR and the text-history policy
on all five dimensions. The deterministic statistics further show fewer exact
or fuzzy repeated searches and fewer response tokens than AgentOCR, consistent
with the matched-state analysis of improved retrieval and stopping decisions.\looseness=-1\par

\FloatBarrier
\section{Prompts and Qualitative Cases}
\label{app:prompts-cases}

\subsection{Agent and diagnostic prompts}

Following AgentOCR~\citep{feng2026agentocr}, we report the visual- and
text-history prompts separately. Offline SFT uses the visual prompts, while the online teacher uses the
corresponding text-history prompt at the student-visited state.

\begin{lstlisting}[style=TaskPrompt,caption={SearchQA visual-history prompt.}]
(*@\promptimage{<image>}@*)

You are an expert agent tasked with answering the given question step-by-step.
Your question: {task_description}

Prior to this step, you have already taken {step_count} step(s).
The image contains the full history:
- Past queries are inside (*@\promptsearch{<search>}@*)...(*@\promptsearch{</search>}@*)
- Past results are inside (*@\promptinfo{<information>}@*)...(*@\promptinfo{</information>}@*)

Now it's your turn to respond for the current step.
You should first conduct a reasoning process. After completing your reasoning, choose only one of the following actions (do not perform both):
(1) If any required knowledge is missing or uncertain, you MUST call a search engine to get more external information using format: (*@\promptsearch{<search>}@*) your query (*@\promptsearch{</search>}@*).
(2) Only if the image/history already provides sufficient, reliable information to answer with high confidence, provide your final answer within (*@\promptanswer{<answer>}@*) (*@\promptanswer{</answer>}@*) tags.
\end{lstlisting}

\begin{lstlisting}[style=TextPrompt,caption={SearchQA text-history prompt.}]
You are an expert agent tasked with answering the given question step-by-step.
Your question: {task_description}

Prior to this step, you have already taken {step_count} step(s). Below is the interaction history, where (*@\promptsearch{<search>}@*)...(*@\promptsearch{</search>}@*) wrapped your past search queries and (*@\promptinfo{<information>}@*)...(*@\promptinfo{</information>}@*) wrapped the corresponding search results. (*@\prompthistory{History:}@*)
(*@\prompthistory{\{memory\_context\}}@*)

Now it's your turn to respond for the current step.
You should first conduct a reasoning process. After completing your reasoning, choose only one of the following actions (do not perform both):
(1) If any required knowledge is missing or uncertain, you MUST call a search engine to get more external information using format: (*@\promptsearch{<search>}@*) your query (*@\promptsearch{</search>}@*).
(2) Only if you have sufficient information to answer the question with high confidence, provide your final answer within (*@\promptanswer{<answer>}@*) (*@\promptanswer{</answer>}@*) tags.
\end{lstlisting}

\begin{lstlisting}[style=TaskPrompt,caption={ALFWorld visual-history prompt.}]
(*@\promptimage{<image>}@*)

You are an expert agent operating in the ALFRED Embodied Environment. Your task is to: {task_description}

Prior to this step, you have already taken {step_count} step(s). The provided image shows the most recent {history_length} observations and the corresponding actions you took.

You are now at step {current_step} and your current textual observation is: {current_observation}
Your admissible actions of the current situation are: [{admissible_actions}].

Now it's your turn to take an action.
You should first reason step-by-step about the current situation. This reasoning process MUST be enclosed within (*@\promptthink{<think>}@*) (*@\promptthink{</think>}@*) tags.
Once you've finished your reasoning, you should choose an admissible action for current step and present it within (*@\promptaction{<action>}@*) (*@\promptaction{</action>}@*) tags.
\end{lstlisting}

\begin{lstlisting}[style=TextPrompt,caption={ALFWorld text-history prompt.}]
You are an expert agent operating in the ALFRED Embodied Environment. Your task is to: {task_description}

Prior to this step, you have already taken {step_count} step(s). Below are the most recent {history_length} observations and the corresponding actions you took: (*@\prompthistory{\{action\_history\}}@*)

You are now at step {current_step} and your current observation is: {current_observation}
Your admissible actions of the current situation are: [{admissible_actions}].

Now it's your turn to take an action.
You should first reason step-by-step about the current situation. This reasoning process MUST be enclosed within (*@\promptthink{<think>}@*) (*@\promptthink{</think>}@*) tags.
Once you've finished your reasoning, you should choose an admissible action for current step and present it within (*@\promptaction{<action>}@*) (*@\promptaction{</action>}@*) tags.
\end{lstlisting}

\begin{lstlisting}[style=TaskPrompt,caption={History QA visual prompt.}]
(*@\promptimage{<image>}@*)

Answer the question using only the interaction history image.
Return only the short answer. Do not explain.
Question: {qa_question}
\end{lstlisting}

The primary rubric request contains three messages: the rubric system
instruction, a dynamically serialized anonymous-trajectory payload, and the
output-format instruction. The template below omits real questions and
trajectory content. In the actual request, history and retrieval fields are
truncated to at most 6,000 characters per field, and model reasoning/output is
truncated to at most 2,500 characters.\looseness=-1\par

\begin{lstlisting}[style=JudgePrompt,caption={Trajectory-judge system prompt.}]
You are evaluating multi-step search-agent trajectories.
The retrieval text shown after each search is the canonical semantic content that was available in that trajectory's history. Do not evaluate OCR transcription. Evaluate whether each anonymous agent used the available information correctly.

Score every anonymous trajectory on five dimensions from 0 to 2:
- evidence_uptake: 0 ignores or contradicts evidence; 1 partial; 2 correctly uses key facts.
- entity_relation_tracking: 0 drifts or confuses; 1 mixed; 2 preserves entities and relations.
- query_progression: 0 redundant or irrelevant; 1 some information gain; 2 targets the precise missing fact.
- stopping_calibration: 0 premature answer or needless searches; 1 defensible but inefficient; 2 searches and answers at the right time.
- answer_grounding: 0 unsupported or contradicted; 1 partially supported; 2 fully supported by retrieved evidence.

For each step, identify whether accumulated evidence before the action was none, partial, or sufficient. Mark useful and redundant search steps. Do not infer model identity, use success labels, or judge final-answer correctness against outside knowledge. Return a complete ranking of all anonymous trajectories. Derived efficiency and modality-gap metrics are computed locally and must not alter these five scores.
\end{lstlisting}

\begin{lstlisting}[style=JudgePrompt,caption={Trajectory-judge input template.}]
{
  "question": "{question}",
  "anonymous_labels": ["A", "B", "C", "D"],
  "trajectories": {
    "{anonymous_label}": {
      "steps": [
        {
          "step_index": {zero_based_step_index},
          "history_before_action": "{canonical_history_before_this_action}",
          "model_reasoning_and_output": "{model_reasoning_and_raw_output}",
          "action": "{postprocessed_search_or_answer_action}",
          "retrieval_text_after_action": "{canonical_retrieval_text_or_empty_string}"
        }
      ],
      "final_answer": "{parsed_final_answer}"
    }
  },
  "allowed_error_tags": [
    "NO_RELEVANT_EVIDENCE_RETRIEVED",
    "EVIDENCE_PRESENT_NOT_USED",
    "ENTITY_DRIFT",
    "RELATION_COMPOSITION_ERROR",
    "REDUNDANT_CONFIRMATION",
    "PREMATURE_ANSWER",
    "LATE_STOP",
    "ANSWER_EXTRACTION_ERROR",
    "CONTRADICTS_EVIDENCE",
    "INVALID_ACTION",
    "UNJUDGEABLE"
  ]
}

The object under "trajectories" is repeated once for every anonymous label.
For Answer actions, "retrieval_text_after_action" is an empty string.
\end{lstlisting}

\begin{lstlisting}[style=JudgePrompt,caption={Trajectory-judge output format.}]
Output-format clarification only. Do not change the semantic rubric, its five dimensions, their meanings, or how the trajectories are compared.
Return exactly one JSON object and no prose outside it.
For every anonymous trajectory:
- scores contains exactly evidence_uptake, entity_relation_tracking, query_progression, stopping_calibration, and answer_grounding; each score is a JSON number from 0 through 2.
- history_utilization_score is the exact numeric sum of those five scores.
- evidence_state_by_step is a JSON array aligned with trajectory steps and uses none, partial, or sufficient.
- first_sufficient_step is a zero-based integer step index or null.
- useful_search_steps and redundant_search_steps are JSON arrays of zero-based integer step indices.
- error_tags is a JSON array using only the documented error tags.
- primary_failure is a string or null; concise_rationale is a non-empty string.
- confidence is a JSON number from 0.0 through 1.0, never a word or percentage.
pairwise.best_history_user is exactly one anonymous label. pairwise.ranking contains every anonymous label exactly once. pairwise.rationale is a non-empty string.
Check all field types, score arithmetic, anonymous labels, and ranking completeness before responding.

Additional step-level schema requirements:
- evidence_state_by_step has exactly one entry per trajectory step and every entry is none, partial, or sufficient.
- first_sufficient_step is null when no step is sufficient; otherwise it is the zero-based index of the first sufficient entry.
- useful_search_steps and redundant_search_steps contain unique zero-based integer indices, are disjoint, reference only actual search actions, and together classify every search action exactly once.
\end{lstlisting}

\begin{figure}[!htbp]
\centering
\begin{minipage}[t]{0.56\linewidth}
\vspace{0pt}
\centering
\includegraphics[width=\linewidth]{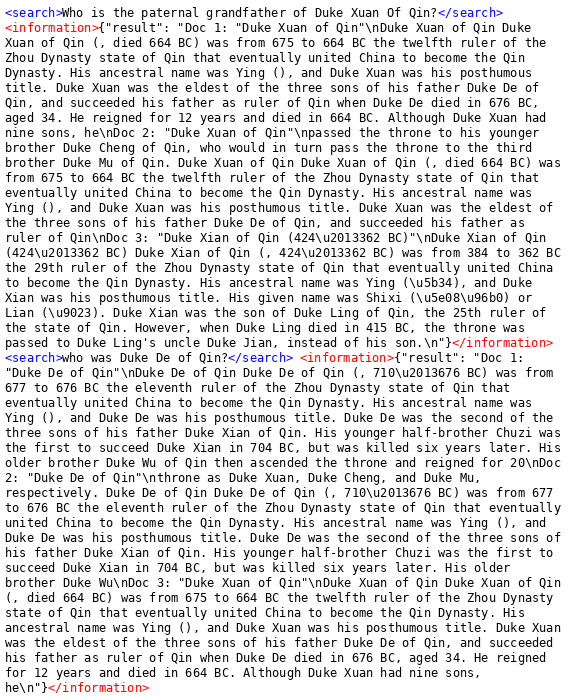}
\par\smallskip
{\small\textbf{(a) SearchQA:} two Search--Information pairs before the third
decision.}

\end{minipage}\hfill
\begin{minipage}[t]{0.41\linewidth}
\vspace{0pt}
\centering
\includegraphics[width=\linewidth]{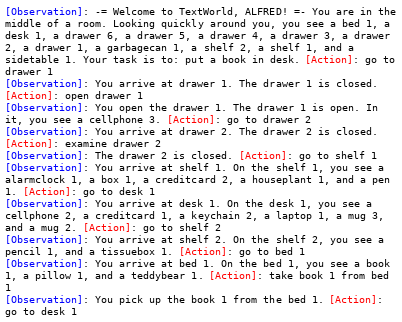}
\par\smallskip
{\small\textbf{(b) ALFWorld:} ten Observation--Action pairs before the eleventh
decision.}

\end{minipage}
\caption{Rendered visual-history examples.}
\label{fig:app-rendered-history-examples}
\end{figure}

Figure~\ref{fig:app-rendered-history-examples} shows representative visual
inputs at two interaction depths using the exact runtime rendering settings,
without additional image compression.

\subsection{Real trajectory cases}

Table~\ref{tab:app-cases} reports representative examples selected from the
finalized rubric set. We show the complete action sequence and the decisive
retrieval excerpts rather than full passages; the judge evaluated the complete
trajectories. Blue marks searches, gold marks retrieved evidence, red marks an
incorrect transition or answer, and green marks the correct answer.

\begin{table}[!htbp]
\centering
\caption{SearchQA trajectory examples.}
\label{tab:app-cases}
\footnotesize
\renewcommand{\arraystretch}{1.08}
\begin{adjustbox}{max width=\linewidth}
\begin{tabular}{>{\raggedright\arraybackslash}p{0.18\textwidth}
                >{\raggedright\arraybackslash}p{0.36\textwidth}
                >{\raggedright\arraybackslash}p{0.36\textwidth}}
\toprule
\textbf{Question and behavior} & \textbf{Comparison policy} & \textbf{\capsmodel} \\
\midrule
\textbf{Corrected entity tracking.}

What is the date of birth of John Jacob Astor VI's mother?

\emph{Required relation:} John Jacob Astor VI $\rightarrow$ mother
$\rightarrow$ birth date. &
\cellcolor{HeaderGray}
\textcolor{PromptSearchTag}{\textbf{Search 1:}}
``Who is John Jacob Astor VI and when was his mother born?''

\textcolor{PromptActionTag}{\textbf{Evidence:}} the results identify John Jacob
Astor VI as the son of \textbf{Madeleine Talmage Force}, but also contain a
distractor about John Jacob Astor VII and Nancy Astor.

\textcolor{PromptInfoTag}{\textbf{Wrong turn:}} AgentOCR follows the distractor
and searches ``When was Viscountess Astor born?''

\textcolor{PromptInfoTag}{\textbf{Answer: ``15 May 1879.''}}

Entity tracking fails despite the relevant mother being present in the first
retrieval. Rubric total: 0. &
\cellcolor{OursBlue}
\textcolor{PromptSearchTag}{\textbf{Search 1:}}
asks directly for John Jacob Astor VI's mother's birth date.

\textcolor{PromptActionTag}{\textbf{Evidence:}} identifies his mother as
\textbf{Madeleine Talmage Force}.

\textcolor{PromptSearchTag}{\textbf{Search 2:}}
``Madeleine Force (mother of John Jacob Astor VI).''

\textcolor{PromptActionTag}{\textbf{Evidence:}} ``Madeleine Talmage Force was
born on \textbf{June 19, 1893}.''

\textcolor{PromptAnswerTag}{\textbf{Answer: ``June 19, 1893.''}}

\capsmodel{} preserves the entity relation and retrieves the missing attribute.
Rubric total: 10. \\
\midrule
\textbf{Reduced redundant search.}

When did Otto II, Prince of Anhalt-Aschersleben's father die?

\emph{Required relation:} Otto II $\rightarrow$ father Otto I $\rightarrow$
death date. &
\cellcolor{HeaderGray}
\textcolor{PromptSearchTag}{\textbf{Search 1:}}
asks when Otto II's father died.

\textcolor{PromptActionTag}{\textbf{Sufficient evidence:}} Otto II was the son of
Otto I, who ``died \textbf{25 June 1304}.''

\textcolor{PromptInfoTag}{\textbf{Redundant continuation:}} Off-SD performs two
more searches, first for Otto I and then for Henry II, Otto I's father.

\textcolor{PromptInfoTag}{\textbf{Answer: ``1266.''}}

It substitutes Henry II's death year for Otto I's after the correct answer was
already available. Rubric total: 0. &
\cellcolor{OursBlue}
\textcolor{PromptSearchTag}{\textbf{Search 1:}}
``Otto II, Prince of Anhalt-Aschersleben's father's death date.''

\textcolor{PromptActionTag}{\textbf{Sufficient evidence:}} the result identifies
the father as Otto I and states that he ``died \textbf{25 June 1304}.''

\textcolor{PromptAnswerTag}{\textbf{Answer: ``25 June 1304.''}}

\capsmodel{} answers at the next step, without redundant retrieval or relation
drift. Rubric total: 10. \\
\bottomrule
\end{tabular}
\end{adjustbox}
\end{table}

The first case isolates entity binding: both initial retrievals contain the
correct mother, but AgentOCR follows a nearby Astor-family distractor, whereas
\capsmodel{} maintains the relation to Madeleine Force. The second isolates
evidence-conditioned stopping: Off-SD continues after retrieving the exact
date and drifts to the preceding generation, while \capsmodel{} answers as soon
as the evidence is sufficient. The improvements therefore reflect how the
visual-history policy uses retrieved evidence, rather than access to different
facts.\looseness=-1\par

\FloatBarrier
\section{Limitations}
\label{app:limitations}

First, rendering choices are important because font size, width, color, and
padding affect the readability and token cost of the visual history. We use the
original AgentOCR rendering parameters directly and do not conduct a
comprehensive ablation over renderer configurations.

Second, offline distillation depends on successful trajectories produced by
the text-history policy. A weaker source policy or insufficient coverage of
successful states may provide less useful initialization, especially for tasks
whose failures are not represented in the retained trajectory set.

Finally, the LLM judge introduces model dependence. The primary rubric has high
coverage and uses blind labels with inverse-probability weighting, but human
evaluation would further strengthen the diagnostic conclusions.\looseness=-1\par

%% file: aaai2027.bib
@article{nakano2021webgpt,
  title={Webgpt: Browser-assisted question-answering with human feedback},
  author={Nakano, Reiichiro and Hilton, Jacob and Balaji, Suchir and Wu, Jeff and Ouyang, Long and Kim, Christina and Hesse, Christopher and Jain, Shantanu and Kosaraju, Vineet and Saunders, William and others},
  journal={arXiv preprint arXiv:2112.09332},
  year={2021}
}

@article{yao2022react,
  title={React: Synergizing reasoning and acting in language models},
  author={Yao, Shunyu and Zhao, Jeffrey and Yu, Dian and Du, Nan and Shafran, Izhak and Narasimhan, Karthik and Cao, Yuan},
  journal={arXiv preprint arXiv:2210.03629},
  year={2022}
}

@article{schick2023toolformer,
  title={Toolformer: Language models can teach themselves to use tools},
  author={Schick, Timo and Dwivedi-Yu, Jane and Dess{\`\i}, Roberto and Raileanu, Roberta and Lomeli, Maria and Hambro, Eric and Zettlemoyer, Luke and Cancedda, Nicola and Scialom, Thomas},
  journal={Advances in neural information processing systems},
  volume={36},
  pages={68539--68551},
  year={2023}
}

@article{kang2025acon,
  title={Acon: Optimizing context compression for long-horizon llm agents},
  author={Kang, Minki and Chen, Wei-Ning and Han, Dongge and Inan, Huseyin A and Wutschitz, Lukas and Chen, Yanzhi and Sim, Robert and Rajmohan, Saravan},
  journal={arXiv preprint arXiv:2510.00615},
  year={2025}
}

@article{lu2026longseeker,
  title={LongSeeker: Elastic Context Orchestration for Long-Horizon Search Agents},
  author={Lu, Yijun and Ye, Rui and Du, Yuwen and Wang, Jiajun and Liu, Songhua and Chen, Siheng},
  journal={arXiv preprint arXiv:2605.05191},
  year={2026}
}

@article{wei2025deepseek,
  title={Deepseek-ocr: Contexts optical compression},
  author={Wei, Haoran and Sun, Yaofeng and Li, Yukun},
  journal={arXiv preprint arXiv:2510.18234},
  year={2025}
}

@article{xing2026vision,
  title={Vision-centric token compression in large language model},
  author={Xing, Ling and Wang, Alex Jinpeng and Yan, Rui and Shu, Xiangbo and Tang, Jinhui},
  journal={Advances in Neural Information Processing Systems},
  volume={38},
  pages={33080--33110},
  year={2026}
}

@inproceedings{cheng2026glyph,
  title={Glyph: Scaling context windows via visual-text compression},
  author={Cheng, Jiale and Liu, Yusen and Zhang, Xinyu and Fei, Yulin and Hong, Wenyi and Lyu, Ruiliang and Wang, Weihan and Su, Zhe and Gu, Xiaotao and Liu, Xiao and others},
  booktitle={Proceedings of the 64th Annual Meeting of the Association for Computational Linguistics (Volume 1: Long Papers)},
  pages={37145--37158},
  year={2026}
}

@article{feng2026agentocr,
  title={AgentOCR: Reimagining Agent History via Optical Self-Compression},
  author={Feng, Lang and Yang, Fuchao and Chen, Feng and Cheng, Xin and Xu, Haiyang and Wan, Zhenglin and Yan, Ming and An, Bo},
  journal={arXiv preprint arXiv:2601.04786},
  year={2026}
}

@article{shao2024deepseekmath,
  title={Deepseekmath: Pushing the limits of mathematical reasoning in open language models},
  author={Shao, Zhihong and Wang, Peiyi and Zhu, Qihao and Xu, Runxin and Song, Junxiao and Bi, Xiao and Zhang, Haowei and Zhang, Mingchuan and Li, YK and Wu, Yang and others},
  journal={arXiv preprint arXiv:2402.03300},
  year={2024}
}

@article{zhao2026self,
  title={Self-Distilled Reasoner: On-Policy Self-Distillation for Large Language Models},
  author={Zhao, Siyan and Xie, Zhihui and Liu, Mengchen and Huang, Jing and Pang, Guan and Chen, Feiyu and Grover, Aditya},
  journal={arXiv preprint arXiv:2601.18734},
  year={2026}
}

@article{li2026rethinking,
  title={Rethinking on-policy distillation of large language models: Phenomenology, mechanism, and recipe},
  author={Li, Yaxuan and Zuo, Yuxin and He, Bingxiang and Zhang, Jinqian and Xiao, Chaojun and Qian, Cheng and Yu, Tianyu and Gao, Huan-ang and Yang, Wenkai and Liu, Zhiyuan and others},
  journal={arXiv preprint arXiv:2604.13016},
  year={2026}
}

@article{lu2026self,
  title={Self-distilled agentic reinforcement learning},
  author={Lu, Zhengxi and Yao, Zhiyuan and Han, Zhuowen and Wang, Zi-Han and Wu, Jinyang and Gu, Qi and Cai, Xunliang and Lu, Weiming and Xiao, Jun and Zhuang, Yueting and others},
  journal={arXiv preprint arXiv:2605.15155},
  year={2026}
}

@article{lu2026skill0,
  title={Skill0: In-context agentic reinforcement learning for skill internalization},
  author={Lu, Zhengxi and Yao, Zhiyuan and Wu, Jinyang and Han, Chengcheng and Gu, Qi and Cai, Xunliang and Lu, Weiming and Xiao, Jun and Zhuang, Yueting and Shen, Yongliang},
  journal={arXiv preprint arXiv:2604.02268},
  year={2026}
}

@article{feng2026group,
  title={Group-in-group policy optimization for llm agent training},
  author={Feng, Lang and Xue, Zhenghai and Liu, Tingcong and An, Bo},
  journal={Advances in Neural Information Processing Systems},
  volume={38},
  pages={46375--46408},
  year={2026}
}

@article{sun2026reading,
  title={Reading, not thinking: Understanding and bridging the modality gap when text becomes pixels in multimodal llms},
  author={Sun, Kaiser and Yuan, Xiaochuang and Liu, Hongjun and Zhao, Chen and Zhang, Cheng and Dredze, Mark and Bai, Fan},
  journal={arXiv preprint arXiv:2603.09095},
  year={2026}
}

@inproceedings{jimenez2024swe,
  title={Swe-bench: Can language models resolve real-world github issues?},
  author={Jimenez, Carlos E and Yang, John and Wettig, Alexander and Yao, Shunyu and Pei, Kexin and Press, Ofir and Narasimhan, Karthik},
  booktitle={International Conference on Learning Representations},
  volume={2024},
  pages={54107--54157},
  year={2024}
}

@article{wang2023voyager,
  title={Voyager: An open-ended embodied agent with large language models},
  author={Wang, Guanzhi and Xie, Yuqi and Jiang, Yunfan and Mandlekar, Ajay and Xiao, Chaowei and Zhu, Yuke and Fan, Linxi and Anandkumar, Anima},
  journal={arXiv preprint arXiv:2305.16291},
  year={2023}
}

@article{shridhar2020alfworld,
  title={Alfworld: Aligning text and embodied environments for interactive learning},
  author={Shridhar, Mohit and Yuan, Xingdi and C{\^o}t{\'e}, Marc-Alexandre and Bisk, Yonatan and Trischler, Adam and Hausknecht, Matthew},
  journal={arXiv preprint arXiv:2010.03768},
  year={2020}
}

@article{jin2025search,
  title={Search-r1: Training llms to reason and leverage search engines with reinforcement learning},
  author={Jin, Bowen and Zeng, Hansi and Yue, Zhenrui and Yoon, Jinsung and Arik, Sercan and Wang, Dong and Zamani, Hamed and Han, Jiawei},
  journal={arXiv preprint arXiv:2503.09516},
  year={2025}
}

@article{rafailov2023direct,
  title={Direct preference optimization: Your language model is secretly a reward model},
  author={Rafailov, Rafael and Sharma, Archit and Mitchell, Eric and Ermon, Stefano and Manning, Christopher D and Finn, Chelsea},
  journal={arXiv preprint arXiv:2305.18290},
  year={2023}
}

@article{sheng2024hybridflow,
  title={Hybridflow: A flexible and efficient rlhf framework},
  author={Sheng, Guangming and Zhang, Chi and Ye, Zilingfeng and Wu, Xibin and Zhang, Wang and Zhang, Ru and Peng, Yanghua and Lin, Haibin and Wu, Chuan},
  journal={arXiv preprint arXiv:2409.19256},
  year={2024}
}

@article{wang2025reinforcement,
  title={Reinforcement learning optimization for large-scale learning: An efficient and user-friendly scaling library},
  author={Wang, Weixun and Xiong, Shaopan and Chen, Gengru and Gao, Wei and Guo, Sheng and He, Yancheng and Huang, Ju and Liu, Jiaheng and Li, Zhendong and Li, Xiaoyang and others},
  journal={arXiv preprint arXiv:2506.06122},
  year={2025}
}

@inproceedings{rust2023language,
  title={Language modelling with pixels},
  author={Rust, Phillip and Lotz, Jonas F and Bugliarello, Emanuele and Salesky, Elizabeth and Lhoneux, Miryam de and Elliott, Desmond},
  booktitle={The Eleventh International Conference on Learning Representations},
  pages={1--32},
  year={2023},
  organization={OpenReview}
}

@article{xing2025see,
  title={See the text: From tokenization to visual reading},
  author={Xing, Ling and Yan, Rui and Wang, Alex Jinpeng and Li, Zechao and Tang, Jinhui},
  journal={arXiv preprint arXiv:2510.18840},
  year={2025}
}

@article{wang2026multimodal,
  title={Multimodal learning with next-token prediction for large multimodal models},
  author={Wang, Xinlong and Cui, Yufeng and Wang, Jinsheng and Zhang, Fan and Wang, Yueze and Zhang, Xiaosong and Luo, Zhengxiong and Sun, Quan and Li, Zhen and Wang, Yuqi and others},
  journal={Nature},
  volume={650},
  number={8101},
  pages={327--333},
  year={2026},
  publisher={Nature Publishing Group UK London}
}

@article{yang2025qwen3,
  title={Qwen3 technical report},
  author={Yang, An and Li, Anfeng and Yang, Baosong and Zhang, Beichen and Hui, Binyuan and Zheng, Bo and Yu, Bowen and Gao, Chang and Huang, Chengen and Lv, Chenxu and others},
  journal={arXiv preprint arXiv:2505.09388},
  year={2025}
}

@inproceedings{gu2024minillm,
  title={Minillm: Knowledge distillation of large language models},
  author={Gu, Yuxian and Dong, Li and Wei, Furu and Huang, Minlie},
  booktitle={The twelfth international conference on learning representations},
  year={2024}
}

@inproceedings{xu2025speculative,
  title={Speculative knowledge distillation: Bridging the teacher-student gap through interleaved sampling},
  author={Xu, Wenda and Han, Rujun and Wang, Zifeng and Le, Long and Madeka, Dhruv and Li, Lei and Wang, William and Agarwal, Rishabh and Lee, Chen-Yu and Pfister, Tomas},
  booktitle={International Conference on Learning Representations},
  volume={2025},
  pages={64616--64646},
  year={2025}
}

@inproceedings{agarwal2024policy,
  title={On-policy distillation of language models: Learning from self-generated mistakes},
  author={Agarwal, Rishabh and Vieillard, Nino and Zhou, Yongchao and Stanczyk, Piotr and Ramos Garea, Sabela and Geist, Matthieu and Bachem, Olivier},
  booktitle={International Conference on Learning Representations},
  volume={2024},
  pages={21246--21263},
  year={2024}
}

@article{xiao2026mimo,
  title={Mimo-v2-flash technical report},
  author={Xiao, Bangjun and Xia, Bingquan and Yang, Bo and Gao, Bofei and Shen, Bowen and Zhang, Chen and He, Chenhong and Lou, Chiheng and Luo, Fuli and Wang, Gang and others},
  journal={arXiv preprint arXiv:2601.02780},
  year={2026}
}

@article{wang2024leveraging,
  title={Leveraging visual tokens for extended text contexts in multi-modal learning},
  author={Wang, Alex Jinpeng and Li, Linjie and Lin, Yiqi and Li, Min and Wang, Lijuan and Shou, Mike Zheng},
  journal={Advances in Neural Information Processing Systems},
  volume={37},
  pages={14325--14348},
  year={2024}
}

@article{shi2026memocr,
  title={MemOCR: Layout-Aware Visual Memory for Efficient Long-Horizon Reasoning},
  author={Shi, Yaorui and Liu, Shugui and Yang, Yu and Mao, Wenyu and Chen, Yuxin and Gu, Qi and Su, Hui and Cai, Xunliang and Wang, Xiang and Zhang, An},
  journal={arXiv preprint arXiv:2601.21468},
  year={2026}
}

@article{chhikara2025mem0,
  title={Mem0: Building production-ready ai agents with scalable long-term memory},
  author={Chhikara, Prateek and Khant, Dev and Aryan, Saket and Singh, Taranjeet and Yadav, Deshraj},
  journal={arXiv preprint arXiv:2504.19413},
  year={2025}
}

@article{xu2026mem,
  title={A-mem: Agentic memory for llm agents},
  author={Xu, Wujiang and Liang, Zujie and Mei, Kai and Gao, Hang and Tan, Juntao and Zhang, Yongfeng},
  journal={Advances in Neural Information Processing Systems},
  volume={38},
  pages={17577--17604},
  year={2026}
}

@article{zhou2025mem1,
  title={Mem1: Learning to synergize memory and reasoning for efficient long-horizon agents},
  author={Zhou, Zijian and Qu, Ao and Wu, Zhaoxuan and Kim, Sunghwan and Prakash, Alok and Rus, Daniela and Zhao, Jinhua and Low, Bryan Kian Hsiang and Liang, Paul Pu},
  journal={arXiv preprint arXiv:2506.15841},
  year={2025}
}

@inproceedings{joshi2017triviaqa,
  title={Triviaqa: A large scale distantly supervised challenge dataset for reading comprehension},
  author={Joshi, Mandar and Choi, Eunsol and Weld, Daniel S and Zettlemoyer, Luke},
  booktitle={Proceedings of the 55th Annual Meeting of the Association for Computational Linguistics (Volume 1: Long Papers)},
  pages={1601--1611},
  year={2017}
}

@inproceedings{mallen2023not,
  title={When not to trust language models: Investigating effectiveness of parametric and non-parametric memories},
  author={Mallen, Alex and Asai, Akari and Zhong, Victor and Das, Rajarshi and Khashabi, Daniel and Hajishirzi, Hannaneh},
  booktitle={Proceedings of the 61st annual meeting of the association for computational linguistics (volume 1: Long papers)},
  pages={9802--9822},
  year={2023}
}

@article{kwiatkowski2019natural,
  title={Natural questions: a benchmark for question answering research},
  author={Kwiatkowski, Tom and Palomaki, Jennimaria and Redfield, Olivia and Collins, Michael and Parikh, Ankur and Alberti, Chris and Epstein, Danielle and Polosukhin, Illia and Devlin, Jacob and Lee, Kenton and others},
  journal={Transactions of the Association for Computational Linguistics},
  volume={7},
  pages={453--466},
  year={2019},
  publisher={MIT Press One Rogers Street, Cambridge, MA 02142-1209, USA journals-info~…}
}

@inproceedings{yang2018hotpotqa,
  title={HotpotQA: A dataset for diverse, explainable multi-hop question answering},
  author={Yang, Zhilin and Qi, Peng and Zhang, Saizheng and Bengio, Yoshua and Cohen, William and Salakhutdinov, Ruslan and Manning, Christopher D},
  booktitle={Proceedings of the 2018 conference on empirical methods in natural language processing},
  pages={2369--2380},
  year={2018}
}

@inproceedings{ho2020constructing,
  title={Constructing a multi-hop qa dataset for comprehensive evaluation of reasoning steps},
  author={Ho, Xanh and Nguyen, Anh-Khoa Duong and Sugawara, Saku and Aizawa, Akiko},
  booktitle={Proceedings of the 28th International Conference on Computational Linguistics},
  pages={6609--6625},
  year={2020}
}

@article{trivedi2022musique,
  title={MuSiQue: Multihop Questions via Single-hop Question Composition},
  author={Trivedi, Harsh and Balasubramanian, Niranjan and Khot, Tushar and Sabharwal, Ashish},
  journal={Transactions of the Association for Computational Linguistics},
  volume={10},
  pages={539--554},
  year={2022},
  publisher={MIT Press One Broadway, 12th Floor, Cambridge, Massachusetts 02142, USA~…}
}

@inproceedings{press2023measuring,
  title={Measuring and narrowing the compositionality gap in language models},
  author={Press, Ofir and Zhang, Muru and Min, Sewon and Schmidt, Ludwig and Smith, Noah A and Lewis, Mike},
  booktitle={Findings of the Association for Computational Linguistics: EMNLP 2023},
  pages={5687--5711},
  year={2023}
}

@misc{bai2025qwen25vltechnicalreport,
      title={Qwen2.5-VL Technical Report}, 
      author={Shuai Bai and Keqin Chen and Xuejing Liu and Jialin Wang and Wenbin Ge and Sibo Song and Kai Dang and Peng Wang and Shijie Wang and Jun Tang and Humen Zhong and Yuanzhi Zhu and Mingkun Yang and Zhaohai Li and Jianqiang Wan and Pengfei Wang and Wei Ding and Zheren Fu and Yiheng Xu and Jiabo Ye and Xi Zhang and Tianbao Xie and Zesen Cheng and Hang Zhang and Zhibo Yang and Haiyang Xu and Junyang Lin},
      year={2025},
      eprint={2502.13923},
      archivePrefix={arXiv},
      primaryClass={cs.CV},
      url={https://arxiv.org/abs/2502.13923}, 
}

@article{deepseekai2026deepseekv4,
  title={Deepseek-v4: Towards highly efficient million-token context intelligence},
  author={Xu, Anyi and Lin, Bangcai and Xue, Bing and Wang, Bingxuan and Xu, Bingzheng and Wu, Bochao and Zhang, Bowei and Lin, Chaofan and Dong, Chen and Ling, Chenchen and others},
  journal={arXiv preprint arXiv:2606.19348},
  year={2026}
}

@article{yuan2026vision,
  title={Vision-opd: Learning to see fine details for multimodal llms via on-policy self-distillation},
  author={Yuan, Qianhao and Lou, Jie and Yu, Xing and Lin, Hongyu and Sun, Le and Han, Xianpei and Lu, Yaojie},
  journal={arXiv preprint arXiv:2605.18740},
  year={2026}
}
